\documentclass{article} % For LaTeX2e
\usepackage{iclr2025_conference,times}

\usepackage{amsmath,amsfonts,bm}

\def\eqref#1{equation~\ref{#1}}
\def\1{\bm{1}}

\DeclareMathAlphabet{\mathsfit}{\encodingdefault}{\sfdefault}{m}{sl}
\SetMathAlphabet{\mathsfit}{bold}{\encodingdefault}{\sfdefault}{bx}{n}

\usepackage{hyperref}
\usepackage{url}

\usepackage{graphicx}
\usepackage{subcaption} % 用于 subfigure 环境

\usepackage{amssymb}
\usepackage{booktabs}
\usepackage{makecell}
\usepackage{subcaption}
\usepackage{enumitem}
\usepackage{multirow}

\usepackage{pifont}
\newcommand{\cmark}{\ding{51}}%
\newcommand{\xmark}{\ding{55}}%

\newcommand{\evidence}[1]{{\color{blue}\textit{#1}}}
\newcommand{\inference}[1]{{\color{green!55!black}#1}}

\title{DetectiveQA: Evaluating Long-Context \\Reasoning on Detective Novels}

\author{Zhe Xu\textsuperscript{1}\thanks{ {} Equal contribution.}, 
Jiasheng Ye\textsuperscript{1}\footnotemark[1],  
Xiaoran Liu\textsuperscript{1}\footnotemark[1], 
Xiangyang Liu\textsuperscript{1}\footnotemark[1], 
Tianxiang Sun\textsuperscript{1},\\
\textbf{Zhigeng Liu\textsuperscript{1}, 
Qipeng Guo\textsuperscript{3}, 
Linlin Li\textsuperscript{2},
Qun Liu\textsuperscript{2},
Xuanjing Huang\textsuperscript{1},
Xipeng Qiu\thanks{ {} Corresponding author.}\textsuperscript{1,3}}\\
\textsuperscript{1}School of Computer Science, Fudan University, \\
\textsuperscript{2}Huawei Noah's Ark Lab, \textsuperscript{3}Shanghai AI Laboratory\\
\texttt{\{zxu24,jsye23,xrliu24\}@m.fudan.edu.cn}, %\texttt{xpqiu@fudan.edu.cn}\\
}

\iclrfinalcopy 
\begin{document}

\maketitle

\begin{abstract}
Recently, significant efforts have been devoted to enhancing the long-context capabilities of Large Language Models (LLMs), particularly in long-context reasoning. To facilitate this research, we propose \textbf{DetectiveQA}, a dataset specifically designed for narrative reasoning within long contexts. We leverage detective novels, averaging over 100k tokens, to create a dataset containing 1200 human-annotated questions in both Chinese and English, each paired with corresponding reference reasoning steps. Furthermore, we introduce a step-wise reasoning metric, which enhances the evaluation of LLMs' reasoning processes. We validate our approach and evaluate the mainstream LLMs, including GPT-4, Claude, and LLaMA, revealing persistent long-context reasoning challenges and demonstrating their evidence-retrieval challenges.
% and data contamination issues. 
Our findings offer valuable insights into the study of long-context reasoning and lay the base for more rigorous evaluations. The evaluation code are publicly accessible via our \href{https://github.com/Phospheneser/DetectiveQA}{GitHub repository}
\end{abstract}

\section{Introduction}\label{sec:intro}

The long-context capabilities of Large Language Models (LLMs) \citep{achiam2023gpt4,anthropic2024claude2,Touvron2023llama2,Sun2024moss,bai2023qwen,2023internlm,zeng2023chatglm}, particularly long-context reasoning\citep{kovcisky2018narrativeqa,sprague2024musrtestinglimitschainofthought,wang2024novelqa,karpinska2024one}, are a key competitive advantage in the current landscape. Recently, OpenAI released the O1 model \citep{openai2024o1}, which not only supports a context length of 128k but can also generate extensive reasoning chains, effectively solving complex reasoning problems in intricate scenarios. As the long-context reasoning capabilities of LLMs improve, there is a growing demand for more challenging and realistic long-context reasoning evaluations\citep{sprague2024musrtestinglimitschainofthought,kuratov2024babilong,zhang2023infinitebench,wang2024novelqa,karpinska2024one}. Among these, narrative reasoning based on detective novels provides a sufficiently realistic and challenging setting\citep{sprague2024musrtestinglimitschainofthought,gu2024detectbenchlargelanguagemodel}. As Ilya, the core developer of GPT-4, has stated, if LLMs possess real understanding when presented with \textit{a detective novel with a complicated plot, a storyline, different characters, numerous events, and mysteries like clues}, LLMs should be able to predict who commits the crime at the last page of the book based on the context\footnote{Ilya Sutskever | GPT4 predicts the next word better | Now upgraded to the more powerful GPT4o \url{https://www.youtube.com/watch?v=1OsHC1vbpc0}}.

\begin{figure}[tb]
    \centering
    \includegraphics[width=\linewidth]{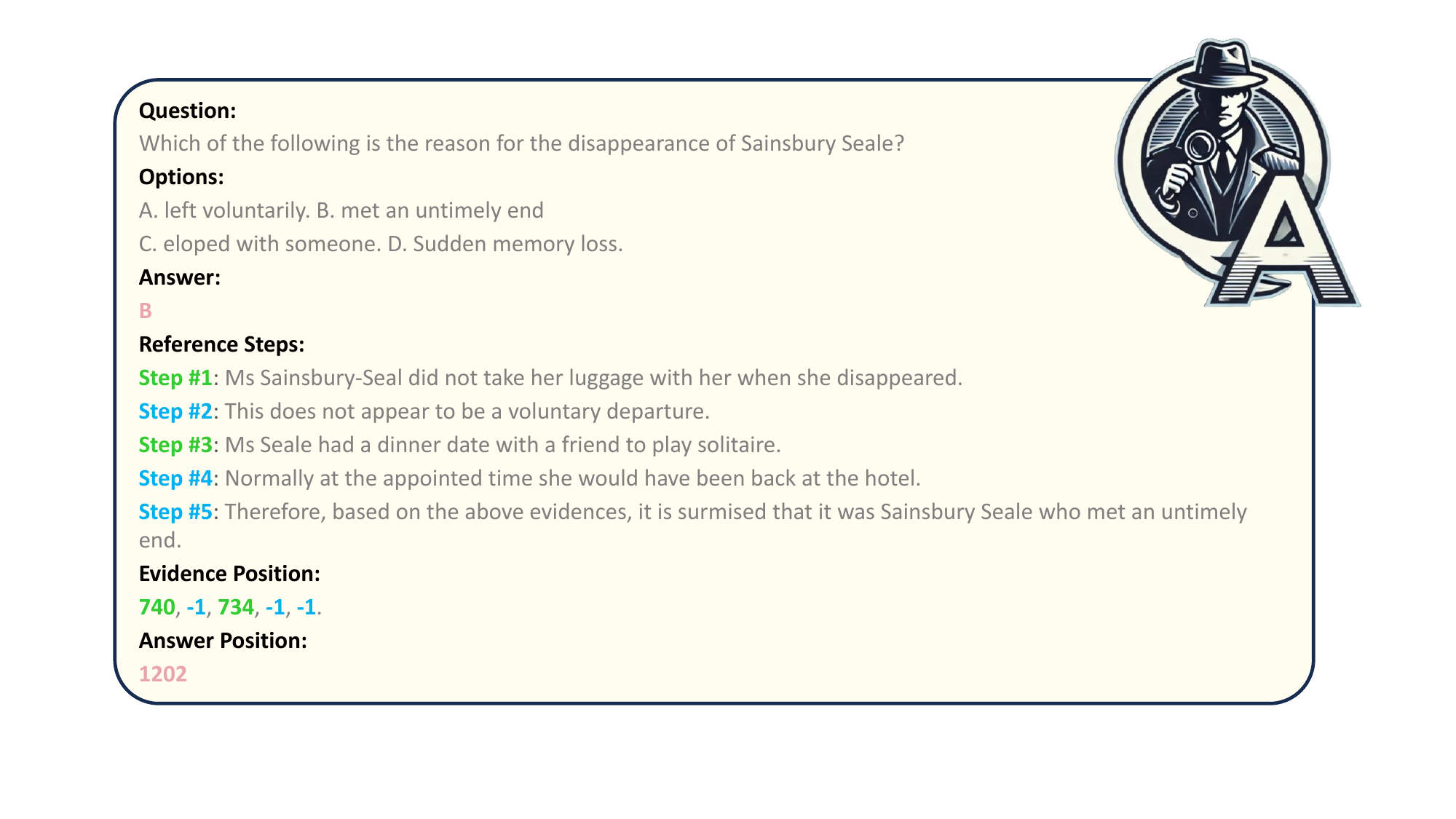}
    \caption{An example of annotation in DetectiveQA. We highlight the \textit{explicit evidence} of reasoning in blue and \textit{implicit evidence} in green. The whole reference steps include both. In contrast, in the \textit{Evidence Position} field, the part corresponding to the explicit evidence will be the paragraph index in the novel, while that corresponding to the implicit evidence will be -1.}
    \vspace*{-5mm}
    \label{fig:detectiveqa}
\end{figure}

Inspired by his words, we propose \textbf{DetectiveQA}, a long-context narrative reasoning dataset with three features. First, DetectiveQA provides \textbf{\textit{detailed annotation information}}. The dataset includes 1200 reasoning questions from English and Chinese detective novels, with an average length exceeding 100k tokens. As shown in Figure~\ref{fig:detectiveqa}, we offer not only the questions, options, and answers but also the reference steps, which are the reference reasoning chains for the question, taken by detectives\citep{wang2024novelqa}. Importantly, these reference steps include the \textit{explicit evidence}, the evidence in the text, and the \textit{implicit evidence}, the inference made by detectives. Our DetectiveQA has the highest average reasoning step number compared with other reasoning datasets as shown in Table~\ref{tab:dataset_comparison}. We also specify the paragraph locations of the evidence and indicate where the detectives give the answer. 

Furthermore, DetectiveQA features \textbf{\textit{step-wise reasoning metric}}. Besides the stable assessment results through multiple-choice questions, we design an LLM-judged metric to evaluate whether LLM's reasoning processes aligns with the steps taken by detectives. This approach offers a more challenging and feasible method for assessing long-context reasoning\citep{an2023leval}. Finally, DetectiveQA can provide an \textbf{\textit{in-depth analysis}} for long-context reasoning. We adjust the evaluation results and the context used during the assessment to explore the LLM's evidence retrieval capability\citep{haystack,li2024needlebench}, the impact of data contamination\citep{li2023loogle,karpinska2024one}, and the differences in reasoning abilities between long and short contexts. To sum up, our contributions can be summarized as follows.

\begin{itemize}
    \item We propose DetectiveQA, a human-annotated evaluation of narrative reasoning in long contexts, averaging over 100k tokens, with 1200 questions in English and Chinese, each paired with reference steps, averaging over 8 steps.
    \item Furthermore, we introduce a step-wise reasoning metric that evaluates the LLM’s reasoning process, which evaluates the evidence retrieval capability of long-context LLMs and, more importantly, reflects the logical coherence of their reasoning process.
    \item We evaluate mainstream LLMs, including GPT-4, Claude, and LLaMA, revealing challenges in long-context reasoning. We also identify data contamination issues and differences in reasoning between long and short contexts. Our findings provide valuable insights for research on long-context LLMs and long-context reasoning assessments.
\end{itemize}

\section{Related Work}\label{sec:related}

\begin{table}[t]
\small
\centering
\caption{Comparison of DetectiveQA with other reasoning datasets. To our knowledge, no previous dataset encompasses all of these qualities. $\thicksim$ denotes datasets that partially qualify for the property.} 
\begin{tabular}{lccccc}
\toprule
Dataset & \textbf{Reasoning} &
\textbf{\makecell{Natural\\Long}} & \textbf{\makecell{Real-\\World}} & \textbf{\makecell{Process\\Evaluation}} & \textbf{\makecell{Reasoning\\Step Num}}
\\
\midrule
NarrativeQA~\citep{kovcisky2018narrativeqa}  & \cmark & \cmark & \cmark & \xmark & 1 \\
HotpotQA ~\citep{yang2018hotpotqa}& \cmark &  \xmark &  \xmark & \xmark & 2 \\
BABILong~\citep{kuratov2024babilong} & \cmark  & \xmark &  \xmark & \xmark & 2.2 \\
NovelQA~\citep{wang2024novelqa} & \cmark  & \cmark &  $\thicksim$ & $\thicksim$ & - \\
NoCha~\citep{karpinska2024one} & \cmark  & \cmark &  $\thicksim$ & \xmark & - \\
MuSR ~\citep{sprague2024musrtestinglimitschainofthought} & \cmark & \xmark & \cmark& \xmark & - \\
DetectBench~\citep{gu2024detectbenchlargelanguagemodel} & \cmark  & \xmark &  \cmark & $\thicksim$ & - \\
\midrule
\textbf{DetectiveQA} (Ours) & \cmark  &  \cmark & \cmark  & \cmark & 8.5 \\
\bottomrule
\end{tabular}

\label{tab:dataset_comparison}
\end{table}

\paragraph{Long-Context Reasoning} Classical long-context benchmark~\citep{bai2023longbench,zhang2023infinitebench,haystack} primarily focuses on tasks like QA, summarization, and retrieval, lacking an evaluation of long-context reasoning in real-world scenarios. Traditional long-context reasoning tasks, such as NarrativeQA~\citep{kovcisky2018narrativeqa}, cover limited clues, while HotpotQA~\citep{yang2018hotpotqa}, despite pioneering multi-hop reasoning, relies on synthetic data that does not provide a realistic context. Recently, NovalQA~\citep{wang2024novelqa} and NoCha~\citep{karpinska2024one} have offered more challenging long-context reasoning evaluations through QA in long novels; however, their question designs are not sufficiently natural and are relatively rare in real-world situations.

Among various genres of novels, detective novels are widely regarded as the most distinctive in terms of reasoning features~\citep{gu2024detectbenchlargelanguagemodel,sprague2024musrtestinglimitschainofthought,del2023true}. Detective reasoning questions can authentically reflect LLM's understanding of context and its reasoning capability in real-world scenarios. Although detective novels have seen successful applications in short-context evaluations, such as MuSR~\citep{sprague2024musrtestinglimitschainofthought} and DetectBench~\citep{gu2024detectbenchlargelanguagemodel}, it has not yet been applied to long-context reasoning evaluation. In response to this gap, we propose DetectiveQA and provide detailed annotation information for classic long-form detective novels.

\vspace*{-3mm}
\paragraph{Reasoning Metrics} To measure the reasoning capability of long-context LLMs, the metric focused on the quality of the reasoning process is necessary. However, the commonly used ROUGE metric~\citep{lin2004rouge} generally fails to do so~\citep{an2023leval}. Therefore, G-Eval utilizes LLMs like GPT-4 to assess the quality of NLG outputs~\citep{liu2023g}, showing a higher correlation with human judgments in summarization and dialogue. Additionally, for mathematical reasoning~\citep{mondorf2024beyond}, ReasonEval~\citep{xia2024evaluating} introduces a method for evaluating the reasoning steps in math problems, emphasizing the validity and redundancy of each step and still using LLMs for automatic assessment. However, these studies have not addressed the reasoning process evaluation in narrative reasoning, particularly in long-context reasoning. For long-context reasoning, the most commonly used metrics remain multiple-choice accuracy~\citep{wang2024novelqa} or output matching~\citep{kuratov2024babilong}. In response to this, we draw inspiration from other process evaluations and, considering the characteristics of detective novels, propose an LLM-judged step-wise reasoning metric. 

\section{Creating DetectiveQA}\label{sec:dataset}

\subsection{Data Source}\label{sec:source}
% A representative data to study language models' ability to handle long contexts is books, among which \textbf{detective novels} are a category that contains intensive reasoning-related content.

\textbf{Detective novels} are valuable for studying language models' ability to handle long contexts due to their reasoning-heavy content. 
Therefore, we consider detective novels as promising candidates to be data sources of our benchmark.
% Nevertheless, we find that a large volume of detective novels take the attractiveness of storytelling in the first place at the sacrifice of the strictness of reasoning processes.
However, many prioritize storytelling over rigorous reasoning. 
Fortunately, we find a group of detective novels categorized as orthodox school~\citep{saito2007orthodox}, which emphasize logical puzzles and provide readers with the same evidence as the detective, making them suitable for our benchmark.
% These novels are dedicated to entertaining readers keen on solving puzzles by ensuring that the reader has the same amount of evidence as the detective in the novel, which is an ideal data source that satisfies our need for rigorous reasoning.
Therefore, we collect orthodox detective novels as sources of long context and use questions related to the puzzles in the novels to test the language models.

% Other considerations of data sources are their lengths and languages. 
% A smooth gradient of difficulty helps to differentiate models at varying levels of proficiency.
% Therefore, we collect orthodox detective novels with lengths ranging from 100k to 250k words.
% Additionally, we only collect the Chinese and English versions given the language background of the researchers and data annotators.
To ensure a smooth gradient of difficulty, we collect novels ranging from 100k to 250k words. We also limit our collection to Chinese and English versions, aligning with the language proficiency of the research team and data annotators.

\begin{figure*}[t] 
\centering
    \includegraphics[width=0.8\textwidth]{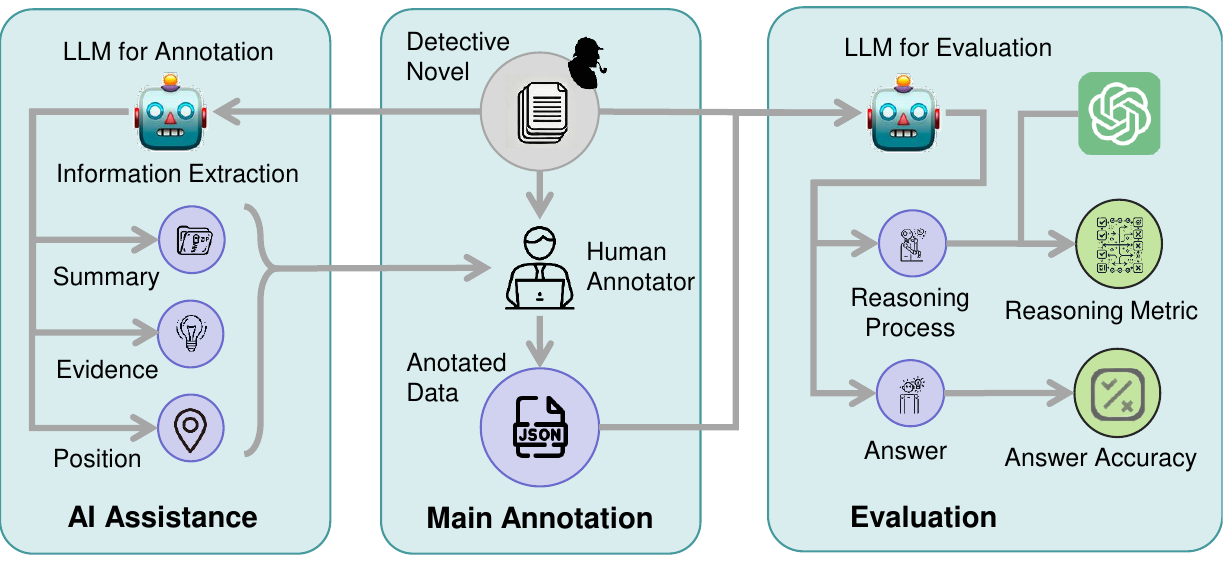}
    \caption{Illustration of DetectiveQA. The center shows the main annotation process, where human annotators annotate reasoning problems based on various information. On the left, AI-assisted information extraction offers summaries to help annotators quickly understand novels and locate key information. The right side, the most critical part, involves evaluating models using DetectiveQA, where reasoning metric and answer accuracy are measured.}
    \label{fig:full process}
\end{figure*}

\subsection{Data Annotation}\label{sec:anno}
% One common method for data annotation is to hire human labor. We could hire workers to read novels, generate questions, and provide reference answers. However, this process is extremely tedious, especially due to the need to read long novels ~\citep{wu2021recursively}. It has been documented that on average, our annotators take approximately 3.5 hours to independently read a 100k-word novel. The high consumption of time and mental effort required from annotators makes it difficult to scale up the dataset. Therefore, we seek an alternative approach to make the data annotation process more efficient.

Data annotation by hiring workers to read long novels~\citep{wu2021recursively} and generate questions is time-consuming and labor-intensive, taking an average of 3.5 hours to read a 100k-word novel, making it challenging to scale, thus prompting the need for a more efficient alternative.

Our solution is to build an agent workflow using existing LLMs with strong long-context capabilities to \textit{assist} human annotators. 
The key insight is that, for a complete detective novel, identifying reasoning questions and their corresponding answers can be viewed as an information extraction task, which is a simpler problem that state-of-the-art LLMs have shown promising performance ~\citep{zhang2023infinitebench}. 
The specific workflow involves the following steps:
% Based on this, we design an agent workflow that decomposes the data annotation process into discrete steps of the information extraction task
% , facilitated by state-of-the-art long-context large language models.

\begin{enumerate}[nosep, label=(\roman*)]
\item \textbf{Novel Comprehension.} We first let the LLM summarize the novel to help the annotator grasp the overall story quickly. We input the entire novel again and ask the model to extract the detective's reasoning within the novel. Allow the annotator to understand the reasoning of the novel better.
% First, we input the complete novel and use the LLMs to extract a summary, which helps the annotator quickly grasp the overall story.
\item \textbf{Question proposition.}
% Then, we use the model to generate questions for each extracted reasoning chain.
Next, Human annotators propose reasoning questions based on the reasoning process extracted.
\item \textbf{Human refinement.}
% To ensure the data quality, the human annotator performs discriminative filtering and refinement on the proposed questions. They also proofread and correct the extracted reasoning based on the corresponding location in the original text.
To aid human annotators in verifying the extracted content, we prepend indices to each paragraph in the novel and instruct the model to output the locations of the extracted reference steps. To ensure data quality, the human annotator performs discriminative filtering and refinement of the extracted reasoning with reference to the corresponding location in the original text.
\end{enumerate}
The above process enhances annotation efficiency while ensuring quality.
Please refer to Appendix~\ref{sec:agent_workflow} for annotation details and Section~\ref{sec:valid} for data validation.
% This approach is designed to improve the efficiency of checking AI-generated annotations. Thanks to manual calibration, we ensure that the annotations are accurate and reasonable. 
% While it still requires a human annotator to proofread and correct the content, this process greatly reduces the overhead of purely manual annotation.
% The validation of the agent workflow-assisted annotated data we will discuss within Section~\ref{sec:valid},  and we will put the detailed agent workflow-assisted annotating process and annotating specification in the 

\section{Evaluating with DetectiveQA}\label{sec:setting}

\subsection{Metric Settings}\label{sec:metric}

In this section, we present the evaluation metrics for DetectiveQA, including multiple-choice accuracy for a stable assessment of reasoning results and the step-wise reasoning metric for a detailed evaluation of the reasoning process.

% \begin{figure}[!tb]
%     \centering
%     \includegraphics[width=0.3\linewidth]{figures/DetectiveQA_1.pdf}
%     \caption{Illustration on the reasoning metrics. The metric evaluates the ability of the models to recall evidence by comparing the models' answers to the reference steps using GPT4.}
%     \label{fig:reasoning}
% \end{figure}

\begin{figure}[!tb]
    \centering
    \begin{subfigure}[b]{0.45\textwidth}
        \includegraphics[width=\linewidth]{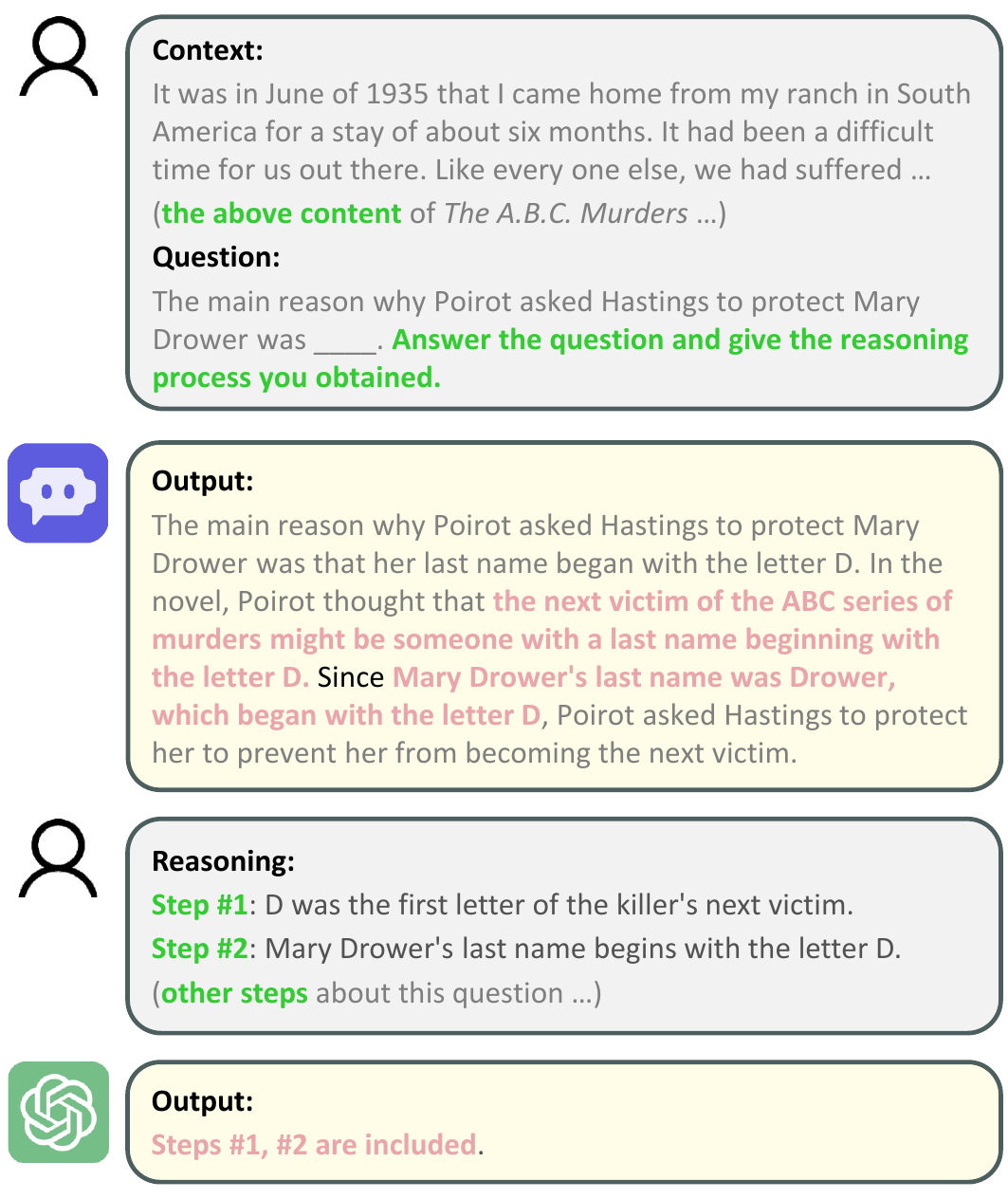}
        \caption{Illustration on the reasoning metrics. The metric evaluates the ability of the models to recall evidence by comparing the models' answers to the reference steps using GPT4.}
        \label{fig:reasoning}
    \end{subfigure}
    \hfill
    \begin{subfigure}[b]{0.45\textwidth}
        \centering
        \begin{subfigure}[b]{\linewidth}
            \centering
            \includegraphics[width=\linewidth]{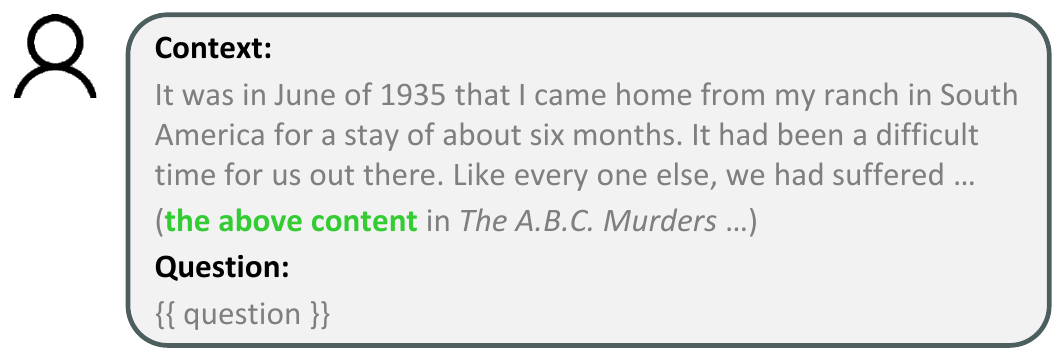}
            \caption{Context+Question}
        \end{subfigure}
        
        \begin{subfigure}[b]{\linewidth}
            \centering
            \includegraphics[width=\linewidth]{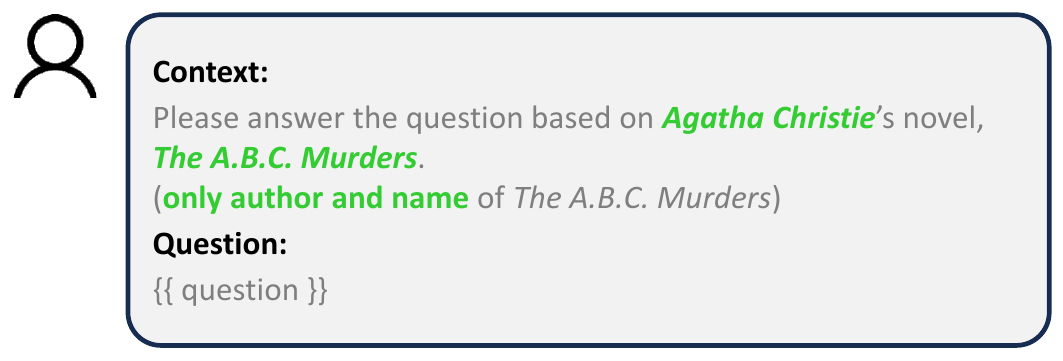}
            \caption{Question-Only}
        \end{subfigure}
        
        \begin{subfigure}[b]{\linewidth}
            \centering
            \includegraphics[width=\linewidth]{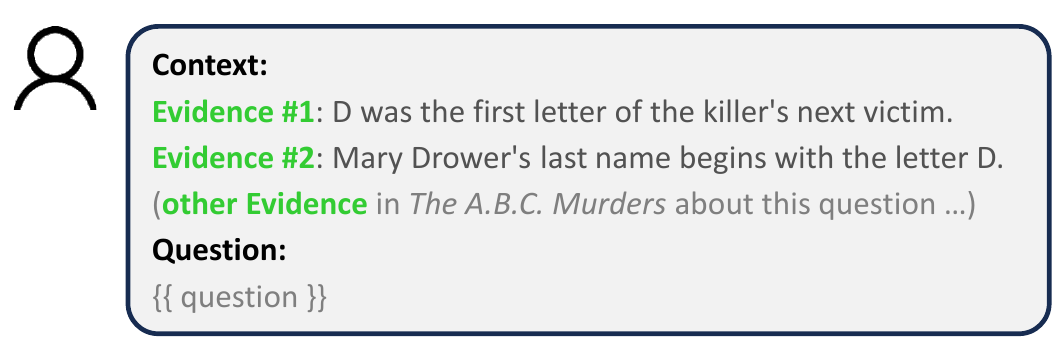}
            \caption{Evidence+Question}
        \end{subfigure}
        \caption{Three context settings for the evaluation on DetectiveQA. \textbf{Context+Question} is the standard setting that requires models to answer questions based on the full long context. \textbf{Question-only} setting prompt without the context, serving for a test of data contamination. \textbf{Evidence+Question} provides the golden evidence instead of the long context, aiming to ablate the reasoning ability of the models without long context.}
        \label{fig:context}
    \end{subfigure}
    \caption{Reasoning metrics with its illustration and three settings.}
    \label{fig:both}
\end{figure}
\vspace*{-3mm}

\paragraph{Multiple-Choice Accuracy} Similar to previous evaluations based on multiple-choice questions~\citep{hendrycks2020MMLU,huang2023CEVAL,an2023leval}, we provide long-context LLM a question with four options and require LLM to output a letter corresponding to the selected option.
At this point, we calculate the percentage of correctly answered questions as the score. 

\paragraph{Step-wise Reasoning Metric} For narrative reasoning in DetectiveQA, simply assessing the answer is insufficient; it is important to measure the logical coherence, namely whether LLM presents sufficient evidence and articulates a complete reasoning process behind the correct answer. However, the automated evaluation of long outputs from LLMs remains challenging\citep{an2023leval}. Fortunately, thanks to the annotation, where each question is linked to corresponding reference steps, we can assess the logical coherence of LLMs' reasoning processes with how many correct reference steps are included, as shown in Figure~\ref{fig:reasoning}. Therefore, we introduce a step-wise reasoning metric, the average score across all questions, to reflect the LLM's reasoning performance on DetectiveQA. 

To access the containment relationship, we use GPT-4 to review and count the reference steps provided in its responses. The specific prompt used is detailed in Appendix~\ref{sec:gpt_score}. Importantly, unlike traditional multi-target information retrieval\citep{haystack,li2024needlebench} or multi-hop reasoning tasks\citep{yang2018hotpotqa,kovcisky2018narrativeqa}, DetectiveQA also evaluates whether the model can provide implicit evidence beyond the evidence present in the context, making it a more challenging and realistic evaluation. We can also visualize evidence retrieval with heatmaps in NIAH\citep{haystack}, which will be discussed in Section~\ref{sec:eval}.

\subsection{Context Settings}\label{sec:context}

% \begin{figure}[!tb]
%     \centering
%     \begin{subfigure}[b]{\linewidth}
%         \centering
%         \includegraphics[width=\linewidth]{figures/DetectiveQA_2a.pdf}
%         \caption{Context+Question}
%     \end{subfigure}
%     \begin{subfigure}[b]{\linewidth}
%         \centering
%         \includegraphics[width=\linewidth]{figures/DetectiveQA_2b2.pdf}
%         \caption{Question-Only}
%     \end{subfigure}
%     \begin{subfigure}[b]{\linewidth}
%         \centering
%         \includegraphics[width=\linewidth]{figures/DetectiveQA_2c.pdf}
%         \caption{Evidence+Question}
%     \end{subfigure}
%     \caption{Three context settings for the evaluation on DetectiveQA. \textbf{Context+Question} is the standard setting that requires models to answer questions based on the full long context. \textbf{Question-only} setting prompt without the context, serving for a test of data contamination. \textbf{Evidence+Question} provides the golden evidence instead of the long context, aiming to ablate the reasoning ability of the models without long context.}
%     \label{fig:context}
% \end{figure}

In addition to two metric settings, we provide three context settings, as shown in Figure~\ref{fig:context}, to analyze different issues in LLMs' long-context reasoning.
\textbf{1.Context+Question} Concatenating the complete context before the answer appears in a detective novel, serving as the basic setting to test LLMs' long-context reasoning capability.
\textbf{2.Question-Only} Only the title and author of the detective novel are provided before the question. This approach addresses potential data contamination issues since our annotated detective novels are classics and may be included in the LLM's pre-training data. By comparing this setting with the previous one, we can assess whether long-context reasoning truly relies on contextual information.
\textbf{3.Evidence+Question} Only relevant evidence from the context is concatenated before the question, transforming long-context reasoning into short-context reasoning. This allows us to compare long-context and short-context reasoning across different LLMs. It is important to note that only the explicit evidence from the context is provided; the implicit evidence from the detective reference steps is not included.

\section{Experiment}\label{sec:exp}
We conduct experiments to validate the data quality of DetectiveQA and study the capability of prominent large language models with it.

\subsection{Validation Results}\label{sec:valid}

\begin{figure}[t]
    \centering
    \includegraphics[width=.4\textwidth]{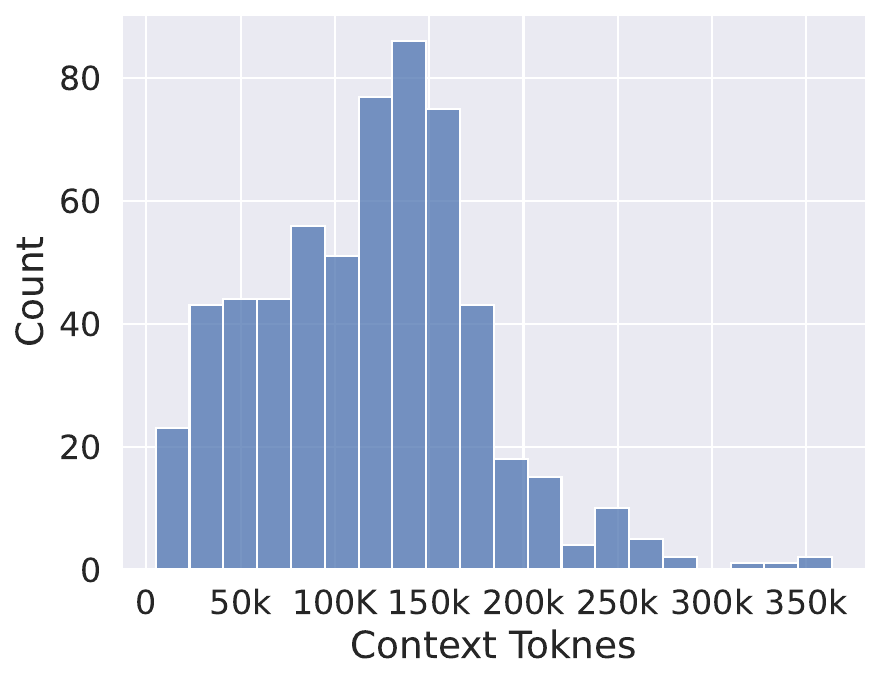}
    \caption{The distribution of the context tokens of samples in DetectiveQA. The novel content for each question is truncated before the answer appears.}
    \label{fig:question-len}
\end{figure}

\paragraph{Statistics.}
We examine the statistics of DetectiveQA to ensure the distribution of both questions and answers satisfies our desiderata.
We confirm that \textbf{(1)} DetectiveQA features long context,  whose context lengths are distributed from 5k to 363k (Figure~\ref{fig:question-len}), with an average of 118k tokens (Table~\ref{tab: all statistic}); \textbf{(2)} The number of clues distributed at each depth exceeded 100, and the exact distribution is shown in the Appendix~\ref{sec:depth distribution}; and \textbf{(3)} the questions are reasoning-intensive, whose answers involve an average of 8.48 reference steps or near 400 tokens~(Table~\ref{tab: all statistic}).
These results support DetectiveQA as a qualified and challenging evaluation for reasoning over long contexts.

\begin{table}[ht]
\caption{
Statistics of evidence and inferences in DetectiveQA.Our dataset also has richer corpus information in terms of responses. We count the lengths in words. And using the GPT4 tokenizer.
}
\centering
\small
\begin{tabular}{lccc} 
\toprule 
 & \textbf{Max.} & \textbf{Min.} & \textbf{Avg.} \\
\midrule 
context tokens & 363k & 5k & 118k \\
% evidence number & 20 & 2 & 6.36 \\
% inference steps & 12 & 1 & 2.34 \\
% evidence length & 812 & 30 & 176.32 \\
% inference length & 493 & 16 & 112.23 \\
% coverage factor & 99.99\% & 0.25\% & 58.54\% \\
reference steps & 27 & 3 & 8.48 \\
reference tokens & 1448 & 96 & 394.95 \\
\bottomrule 
\end{tabular}

\label{tab: all statistic}
\end{table}

\paragraph{Validation on data annotations.}
Three authors of this work double-check the data to validate the annotation quality.
Specifically, we sample annotated data from 10 novels and the three authors inspect (1) whether the \textbf{questions} are reasoning questions whose answers can be derived from the novel; (2) whether the \textbf{reference steps} are consistent with the original novel and all the evidence exists in the novel; and (3) whether the annotated \textbf{answers} are correct.
The results in Table~\ref{tab: anno valid} indicate that the quality of almost all the data satisfies our requirement, thus supporting the validity of the outcome dataset. More supporting evidence will be provided in the Appendix~\ref{sec:more_valid}.

\begin{table}[h]
\centering
\small
\caption{We asked three authors to score each of the ten novels using agent workflow to accelerate the labeling of questions in terms of question reliability, answer accuracy, and reasonableness of the reference step, and used a voting mechanism to arrive at a final score, and computed the Jaccard correlation between this result and the full 1-sequence.
}
\begin{tabular}{lccc} 
\toprule \textbf{Validation part} 
& \textbf{question} & \textbf{reasoning} & \textbf{answer}  \\
\midrule 
Jaccard similarity & 0.97 & 0.91 & 0.94\\
\bottomrule 
\end{tabular}

\label{tab: anno valid}
\end{table}

\paragraph{Validation on metrics.}
We verify the reliability of LLM as a judge for the step-wise reasoning metric.
Specifically, we manually annotate the reasoning process of 100 answers and calculate the correlation between the judgment from the human judges and GPT4. 
As shown in Table~\ref{tab: agreement}, GPT4 shows high agreement with human judges.
In comparison, Rouge, an n-gram-based automatic metric for content recall, despite being much cheaper, fails to correlate with human evaluation well (Table~\ref{tab: CC}). 
Therefore, we consider it appropriate to apply GPT4 as an automatic evaluation of the reasoning process.

\begin{table}[ht]
\centering
\small
\caption{
Using whether evidence is present in the chain of reasoning as a discrete judgment problem, we analyze the consistency of human judgments and GPT4 judgments using human judgments as the gold standard.
}
\begin{tabular}{lcc} 
\toprule 
\textbf{Human vs GPT4} & accuracy & kappa\\
\midrule 
\textbf{Agreement} &  0.92 &  0.83\\ 
\bottomrule 
\end{tabular}

\label{tab: agreement}
\end{table}

Overall, the above validation on both data and evaluation metrics ensures DetectiveQA is an evaluation that satisfies our needs for long-context reasoning ability for large language models. 
We then apply it to evaluate mainstream models and study their capabilities.

\begin{table}[t]
\centering
\small
\caption{
Correlation coefficient table. We randomly selected 100 reasoning processes generated by the model and compared them with human-labeled reference steps. For each process, we computed Rouge-1, Rouge-2, and Rouge-L scores, along with human and GPT ratings. The scores for the 100 reasoning processes, based on these four evaluation metrics, were then correlated with the human ratings to determine the correlation coefficients.
}
\begin{tabular}{lcccc} 
\toprule 
& \textbf{Ours} & \textbf{Rouge-1} & \textbf{Rouge-2} & \textbf{Rouge-L} \\
\midrule 
Corr. & 0.91 & 0.52 & 0.61 & 0.58 \\
\bottomrule 
\end{tabular}

\label{tab: CC}
\end{table}

\subsection{Evaluation Results}\label{sec:eval}

\paragraph{Models.} 
We evaluate both closed-source and open-source LLMs featuring long-context capability.
For closed-source models, we include  GPT4-1106-preview-128k, 
OpenAI-O1-mini-128k~\citep{openai2024o1}, 
Claude3-opus-20240229-200k~\citep{anthropic2024claude3}, and KimiChat-200k.
For open-source models, we evaluate LLaMA3.1-8B-Instruct-128k~\citep{dubey2024llama}, ChatGLM3-6B-128k~\citep{zeng2022glm}, GLM4-9B-chat-1M~\citep{zeng2022glm}, IntermLM2-7B-chat-200k~\citep{2023internlm}, InternLM2.5-7B-chat-1M~\citep{InternLM25} and Qwen2.5-7B-Instruct-128k~\citep{qwen25}.

\paragraph{Main results.}
We first report the multiple-choice accuracy and reasoning metric under the \textit{Question+Context} setting in Table~\ref{tab:main_results}.

Comparing the metrics, we find the multiple-choice accuracy and reasoning metrics show consistent trends across models, while gaps in numbers exist.
% Although both metrics , reasoning metrics are much lower in number. 
This indicates a large number of questions are answered without a perfect reasoning chain, further highlighting the necessity to perform fine-grained evaluation on reference steps to study long-context reasoning capabilities.

\begin{table*}[tbh]
\centering
\small
\caption{Win rate was calculated for model responses based on the Question Only setting and the Question+Context setting, and G.M. is the geometric mean of the answer accuracy and reasoning scores.}
\begin{tabular}{lccccccc}  
\toprule
\multirow{2}{*}{\textbf{Models}} & \multicolumn{3}{c}{\textbf{Question+Context}} & \multicolumn{3}{c}{\textbf{Question-Only}} & \multirow{2}{*}{\textbf{Win Rate}}\\ 
& Answer & Reasoning & G.M. & Answer & Reasoning & G.M. & \\
\midrule
GPT-4-1106-preview-128k & 73.99 & 27.43 & 45.05 & 43.16 & 10.99 & 21.77 & 84.34 \\ 
OpenAI-O1-mini-128k  & 60.83 & 23.80 & 38.05 & 41.67 & 11.64 & \textbf{22.03} & 70.65\\
KimiChat-200k & 64.13 & 27.79 & 42.21 &	\textbf{45.07} & 9.64 & 20.84 & 67.27 \\
Claude3-Opus-200k & \textbf{81.95} & \textbf{37.33} & \textbf{55.30} & 23.43 &	\textbf{16.22} & 19.49	& \textbf{94.61} \\ 
\midrule
LLaMA3.1-8B-Instruct-128k & 28.17 & 21.15 & 24.41 & 39.42 & 8.08 & 17.84 & 69.44 \\
GLM3-6B-128k & 40.58 & 22.08 & 33.63 & 33.63 & 7.16 & 15.51 & 63.47 \\
GLM4-9B-chat-1M & 59.00 & \textbf{24.07} &  \textbf{37.68} & 40.33 & 8.06 & 18.03 & 74.02 \\
Qwen2.5-7B-Instruct-128k & \textbf{61.75} & 21.16 & 36.15 & \textbf{40.58}  & 9.09 & 19.21 & 76.86 \\
InternLM2-7B-chat-200k & 57.95 & 23.94 &37.24 & 36.97& \textbf{12.65} & \textbf{21.62} &	\textbf{81.69}\\ 
InternLM2.5-7B-chat-1M & 60.92 & 22.45 & 36.98 & 39.17 & 7.76 & 17.44 & 77.99 \\
\bottomrule
\end{tabular} 

\label{tab:main_results}
\end{table*}  

Comparing across models, results indicate obvious discrepancies, which we summarize as follows.
\begin{itemize}[nosep]
    \item \textbf{Open-source models still lag behind esteem closed-source models.} While open-source models have made significant progress in recent development, claiming to approach state-of-the-art closed-source models~\citep {dubey2024llama}, the performance gaps still exist in long-context reasoning capabilities.
    \item \textbf{Distinction also exists among closed-source models.} Despite being known as a strong reasoner, OpenaAI-O1-mini-128k does not show distinctively superior performance. Instead, Claude3 performed the best among the closed-source models. % One of the special phenomena we observed was that Claude3 would refuse to answer questions in the Question-only setting.
    % \textcolor{red}{TODO:O1...} In particular, Claude3-Opus-200k are distinct in reasoning metric...
    \item \textbf{Most open-source models perform on par,} while Llama3.1 lags behind others. Our subsequent analysis (Section~\ref{sec:ablation}) attributes this to its failure on over 100k-long contexts. 
\end{itemize}

\begin{figure}[t]
    \centering
    \begin{subfigure}[b]{0.45\textwidth}
        \includegraphics[width=\textwidth]{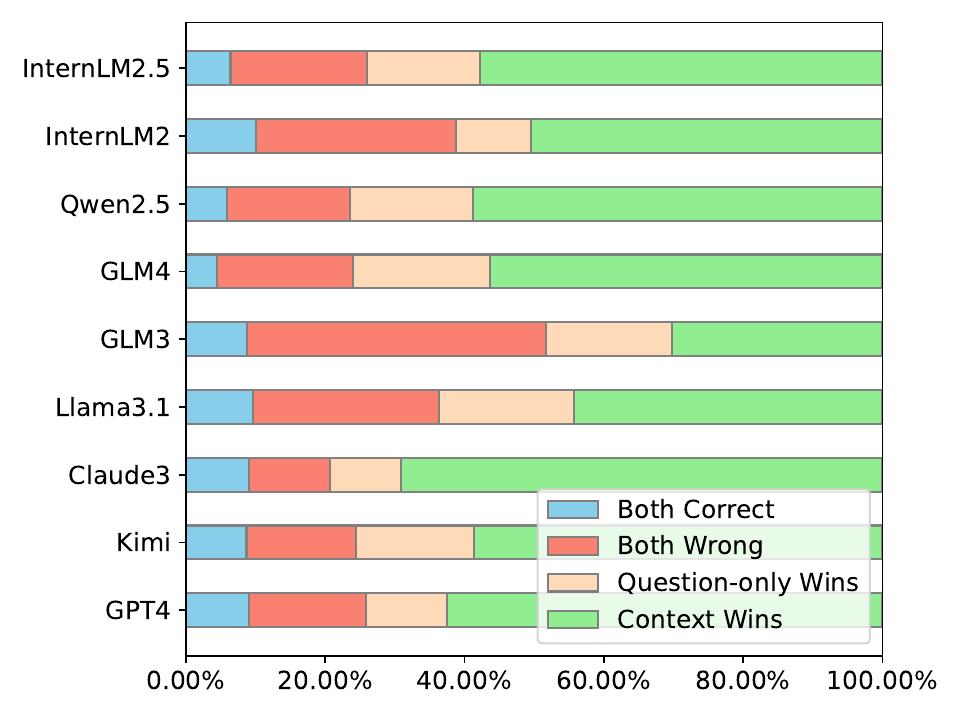}
        \caption{Stacked bar charts for analyzing data contamination. The figure contrasts model perf. in Q-only vs. C+Q modes using win rate (Sec.~\ref{sec:eval}). If both settings yield incorrect answers, they’re not compared, categorized as “both lose” for calc.}
        \label{fig:CvsQ}
    \end{subfigure}
    \hfill
    \begin{subfigure}[b]{0.45\textwidth}
        \includegraphics[width=\textwidth]{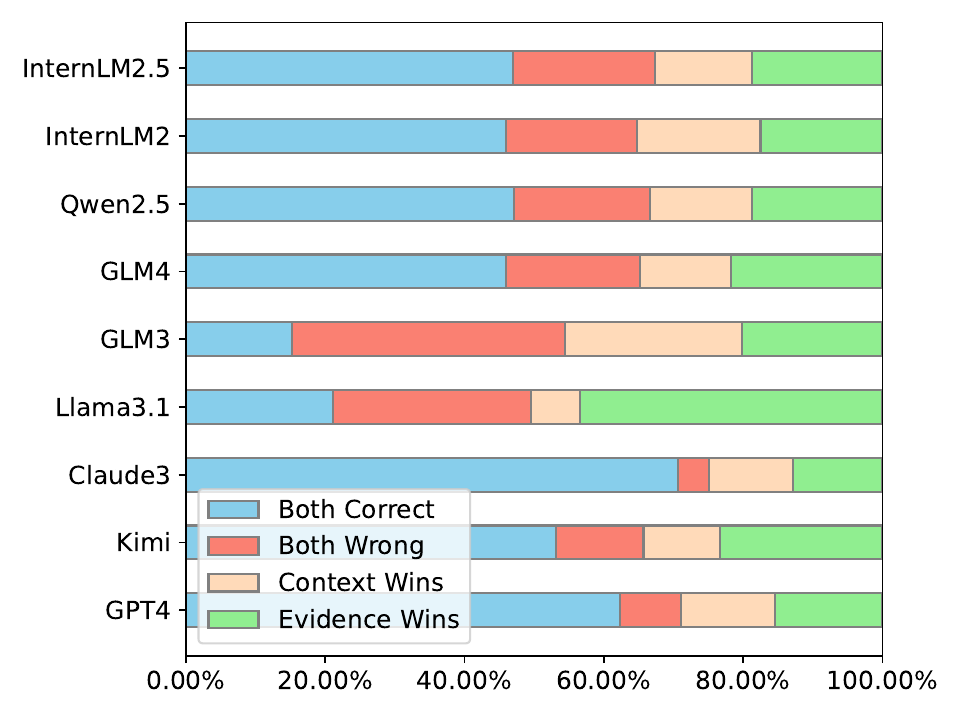}
        \caption{Stacked bar charts for reasoning analysis. The figure shows model perf. in evidence+Q and context+Q settings, focusing only on answer accuracy, as reasoning score comparison is unfair to context+Q.}
        \label{fig:CvsE}
    \end{subfigure}
    \caption{Figures of two different analyses.}
    \label{fig:both2}
\end{figure}

\paragraph{Analysis on data contamination.}
The use of detective novels may raise concerns about data contamination issues, which
we investigate the potential impact through the question-only setting.
The main idea is that the model is able to answer the question correctly even without the context if data contamination occurs.
Therefore, we perform a question-wise comparison between the \textit{Question-only} and \textit{Question+Context} settings.
We summarize the results in terms of win rate\footnote{
% \textcolor{red}{TODO: explain our win rate is calculated here} 
Win rate is defined as the score comparison of each question under two different settings, with scores comprising accuracy (0 or 1) and step-wise reasoning metric (0-1), where the higher score wins.
} in Figure~\ref{fig:CvsQ}.
The results suggest the data contamination issues are mild.
The question-only setting merely wins on a small proportion of questions, and the proportions are similar for different models.
Besides, as shown in Table~\ref{tab:main_results}, our main results on \textit{question+context} settings correlate with the win rate.
Therefore, we consider the data contamination is not severe and does not affect the validity of our evaluation with DetectiveQA.

\begin{figure*}[tbh]
    \centering
    % 第一行子图
    \begin{subfigure}[b]{0.32\textwidth}
        \centering
        \includegraphics[width=1\textwidth]{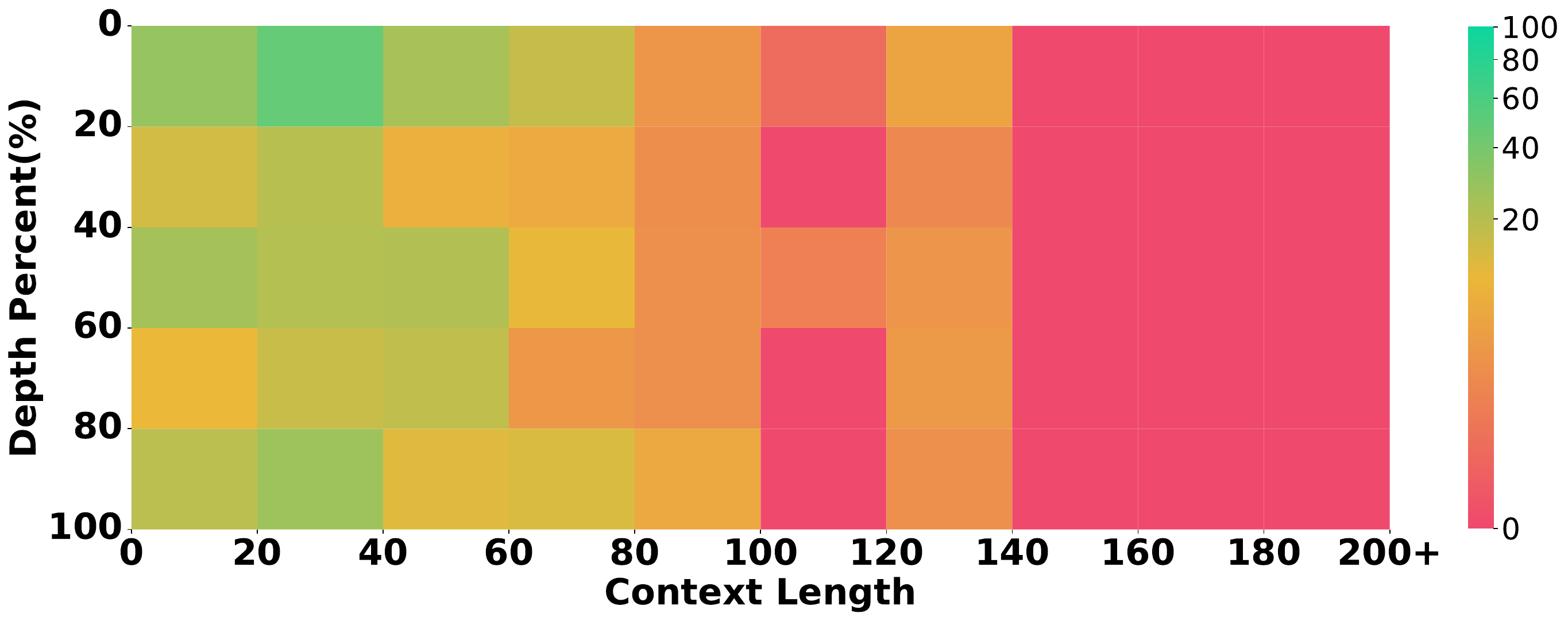}
        \caption{Llama3.1}
    \end{subfigure}
    \hfill
    \begin{subfigure}[b]{0.32\textwidth}
        \centering
        \includegraphics[width=1\textwidth]{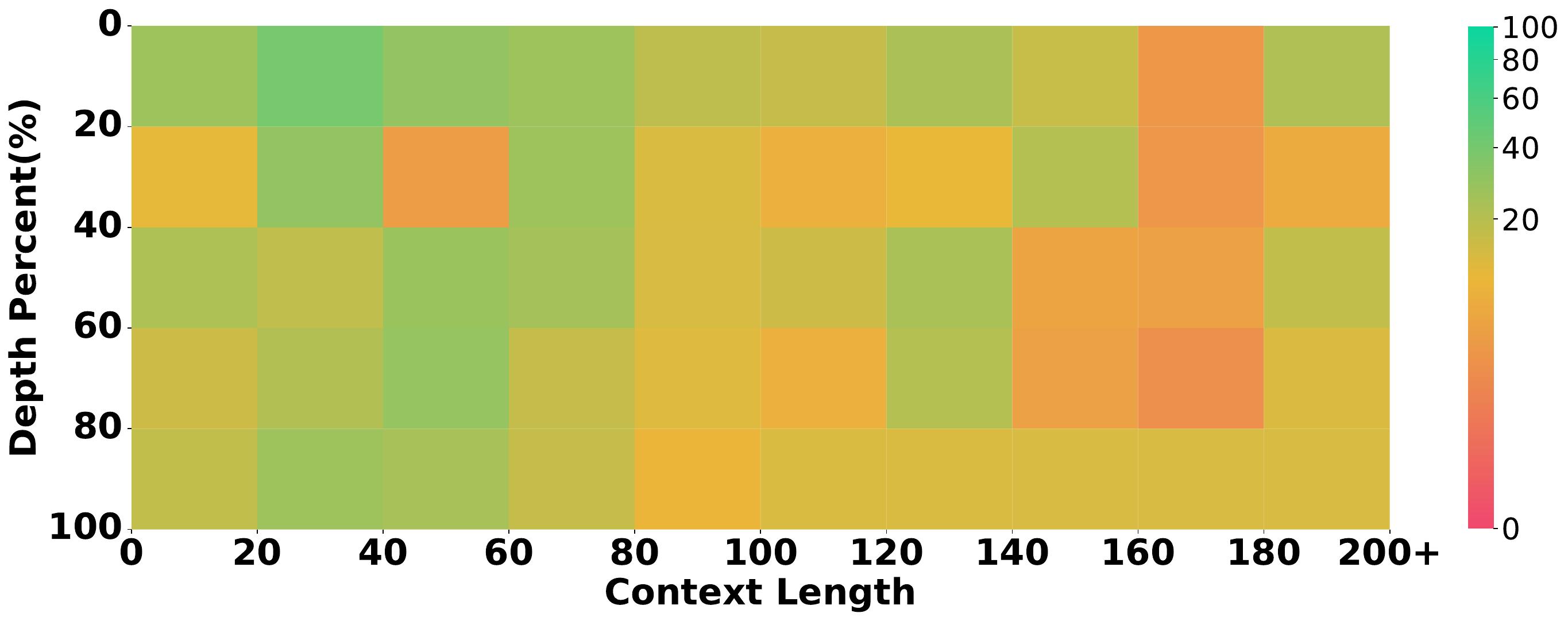}
        \caption{Qwen2.5}
    \end{subfigure}
    \hfill 
    \begin{subfigure}[b]{0.32\textwidth}
        \centering
        \includegraphics[width=1\textwidth]{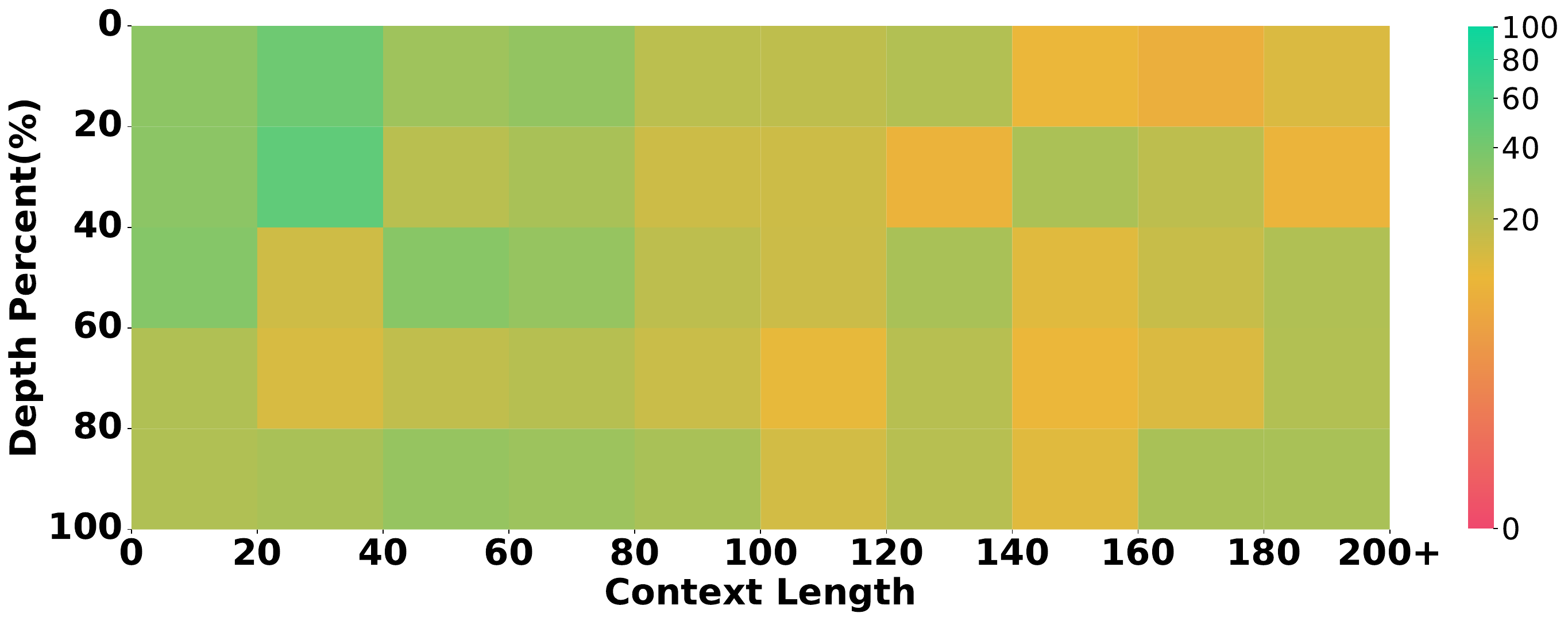}
        \caption{GLM4}
    \end{subfigure}
    
    % 第二行子图
    \begin{subfigure}[b]{0.32\textwidth}
        \centering
        \includegraphics[width=1\textwidth]{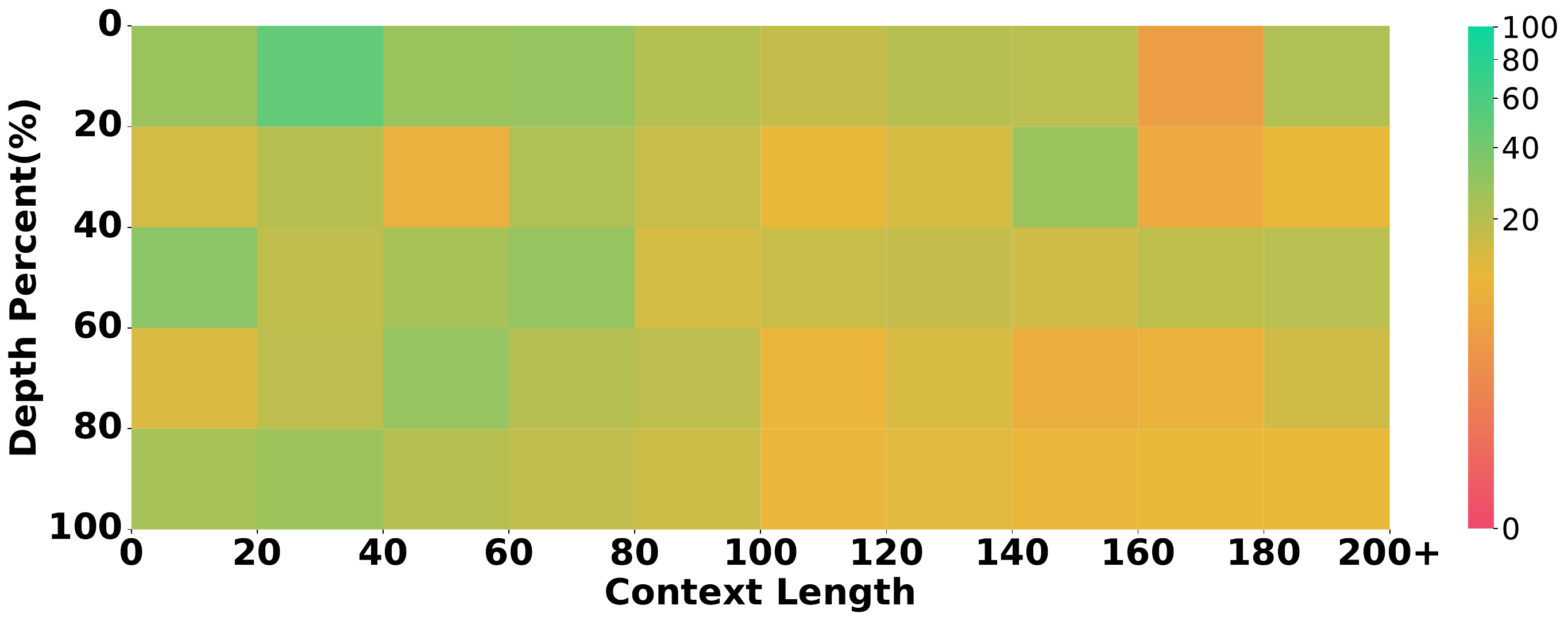}
        \caption{InternLM2.5}
    \end{subfigure}
    \hfill
    \begin{subfigure}[b]{0.32\textwidth}
        \centering
        \includegraphics[width=1\textwidth]{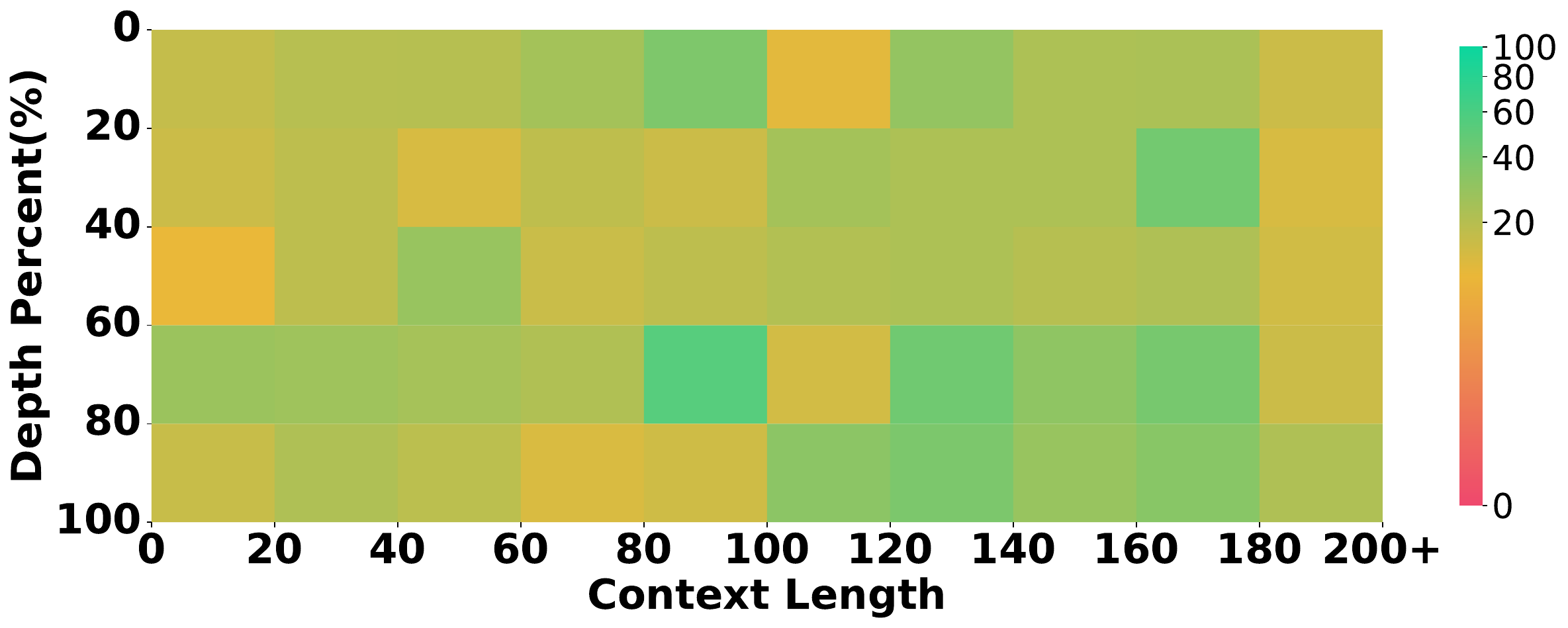}
        \caption{GPT4}
    \end{subfigure}
    \hfill
    \begin{subfigure}[b]{0.32\textwidth}
        \centering
        \includegraphics[width=1\textwidth]{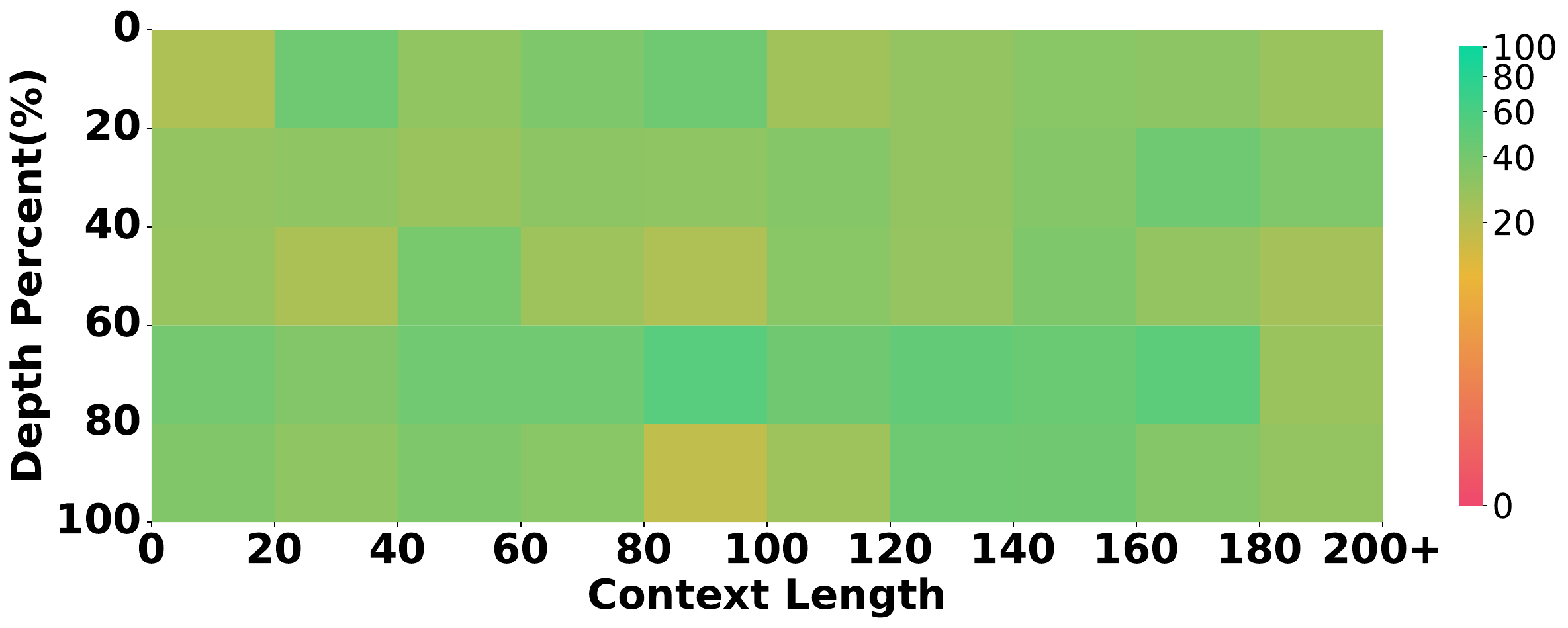}
        \caption{Claude3}
    \end{subfigure}

    \caption{Multi-needle-in-a-haystack test results for different models. We treat each clue in the reference steps found in the article as a "needle" and determine whether the needle is detected by checking if it is included in the model's reasoning process. We define "depth" as the percentage of the problem's total character count where the evidence appears far from the beginning of the document. Our analysis focuses on recall based on varying context lengths and clue depths.}
    \label{fig:multi-nihs}
\end{figure*}

\subsection{Identifying Performance Bottlenecks by Ablating  Long Context and Reasoning}
\label{sec:ablation}
DetectiveQA evaluates the intersection of long-context processing and reasoning, two crucial abilities for large language models, simultaneously.
To help identify the performance bottlenecks and provide insight to help improve model capability, we disentangle the effect of the two capabilities.

\subsubsection{Analysis on evidence retrieval over long context.}
\label{sec:ablate-long}
To understand whether the evaluated models are proficient in long-context processing, we study the models' success rate in retrieving evidence in different positions over the context of different lengths, inspired by the multi-needle in a haystack task~\citep{li2024needlebench}.

Results are in figure~\ref{fig:multi-nihs}, which unveils that some models underperform others due to their limitations in long-context processing. 
For instance, the overall performance of Llama3.1 in our main setting lags behind others. 
Correspondingly, Figure~\ref{fig:multi-nihs} shows that Llama3.1 almost fails to retrieve any clues when the context length exceeds 100k tokens, while the others still show a descent recall rate.
This showcases our annotation on evidence and their position being helpful to ablate the long-context abilities in long-context reasoning.

\subsubsection{Analysis on reasoning capability.}
For how reasoning affects model performance on DetectiveQA, we study the comparison between \textit{Question+Context} and \textit{Evidence+Context} settings.
The latter provides golden evidence instead of the long context, thus eliminating the influence of the long-context capability of the models. 
Similar to the comparison between \textit{question-only} and \textit{Question+Context} settings, we again conduct question-wise comparisons and summarize the win rates in the two settings, shown in figure~\ref{fig:CvsE}.

We summarize our findings from different types of answers, respectively.
The \textbf{"evidence win"} takes up a number of answers for all the models, indicating the need to enhance long-context capability for all models.
Notably, the answers from Llama 3.1 take up a large proportion in this part, aligning with our previous findings on its limitation in long-context processing (Section~\ref{sec:ablate-long}).
Additionally, comparing the proportions of being \textbf{"both wrong"}, we find GLM3 stands out to have more answers belonging to this type, implying its less advanced reasoning ability.
We also notice there exist a non-neglectable number of cases in \textbf{"context win"}, possibly due to that dropping most of the context may harm the understanding of the story thus harming performance\footnote{This may relate to the comparison between retrieval-based methods~\citep{xu2023ragvslong} and context-based methods for handling massive information. We leave this for future study. }.

\vspace{-3mm}
\section{Conclusion}\label{sec:conclusion}
\vspace{-0.5mm}
% \vspace{-1÷mm}
% \textcolor{red}{TODO: write conclusion!}
We introduced DetectiveQA to test the models' ability to reason narratively over long contexts, the first benchmark for narrative reasoning with an average context length of 100k. We challenged the models' ability to reason over long texts as well as narrative reasoning using detective novels, the real-world texts. For each model, our test gives two scores (answer accuracy and reasoning score) in three settings. With a rich experimental setup, we can deeply analyze the performance of the model and find that the current model still faces challenges in long text comprehension, information extraction and narrative reasoning.
We hope that our dataset will facilitate future improvements in model reasoning ability, leading to more robust AI applications and the highest machine intelligence.

% \section*{Limitations}
% Our dataset only serves as an evaluation benchmark on long-context reasoning ability, while how to improve the model capability remains an open question. Meanwhile, our benchmark contains only data from detective novels and mainly serves narrative reasoning. More diverse scenarios can be included in the future. 

% \section*{Ethics Statement}
% We are committed to ensuring that DetectiveQA is used only for academic and scientific purposes, and therefore we have rigorously copyright-checked all of the reasoning novels used in Detective's annotations to ensure that the individual novels are not designed to create copyright problems in non-commercial areas. Through these screening tools, we aim to respect the principle of ‘fair use’ under copyright protection and ensure that our project navigates within legal and ethical boundaries in a responsible manner.

\section*{Limitations}
Our dataset only serves as an evaluation benchmark on long-context reasoning ability, while how to improve the model capability remains an open question. Meanwhile, our benchmark contains only data from detective novels and mainly serves narrative reasoning. More diverse scenarios can be included in the future. 

\section*{Ethics Statement}
We are committed to ensuring that DetectiveQA is used only for academic and scientific purposes, and therefore we have rigorously copyright-checked all of the reasoning novels used in Detective's annotations to ensure that the individual novels are not designed to create copyright problems in non-commercial areas. Through these screening tools, we aim to respect the principle of ‘fair use’ under copyright protection and ensure that our project navigates within legal and ethical boundaries in a responsible manner.

\bibliography{iclr2025_conference}

\begin{thebibliography}{36}
\providecommand{\natexlab}[1]{#1}
\providecommand{\url}[1]{\texttt{#1}}
\expandafter\ifx\csname urlstyle\endcsname\relax
  \providecommand{\doi}[1]{doi: #1}\else
  \providecommand{\doi}{doi: \begingroup \urlstyle{rm}\Url}\fi

\bibitem[An et~al.(2023)An, Gong, Zhong, Li, Zhang, Kong, and Qiu]{an2023leval}
Chenxin An, Shansan Gong, Ming Zhong, Mukai Li, Jun Zhang, Lingpeng Kong, and Xipeng Qiu.
\newblock L-eval: Instituting standardized evaluation for long context language models.
\newblock \emph{CoRR}, abs/2307.11088, 2023.
\newblock \doi{10.48550/ARXIV.2307.11088}.
\newblock URL \url{https://doi.org/10.48550/arXiv.2307.11088}.

\bibitem[Anthropic(2024{\natexlab{a}})]{anthropic2024claude2}
Anthropic.
\newblock Model card and evaluations for claude models, 2024{\natexlab{a}}.
\newblock URL \url{https://www-cdn.anthropic.com/bd2a28d2535bfb0494cc8e2a3bf135d2e7523226/Model-Card-Claude-2.pdf}.

\bibitem[Anthropic(2024{\natexlab{b}})]{anthropic2024claude3}
Anthropic.
\newblock Introducing the next generation of claude, 2024{\natexlab{b}}.
\newblock URL \url{https://www.anthropic.com/news/claude-3-family}.

\bibitem[Bai et~al.(2023{\natexlab{a}})Bai, Bai, Chu, Cui, Dang, Deng, Fan, Ge, Han, Huang, Hui, Ji, Li, Lin, Lin, Liu, Liu, Lu, Lu, Ma, Men, Ren, Ren, Tan, Tan, Tu, Wang, Wang, Wang, Wu, Xu, Xu, Yang, Yang, Yang, Yang, Yao, Yu, Yuan, Yuan, Zhang, Zhang, Zhang, Zhang, Zhou, Zhou, Zhou, and Zhu]{bai2023qwen}
Jinze Bai, Shuai Bai, Yunfei Chu, Zeyu Cui, Kai Dang, Xiaodong Deng, Yang Fan, Wenbin Ge, Yu~Han, Fei Huang, Binyuan Hui, Luo Ji, Mei Li, Junyang Lin, Runji Lin, Dayiheng Liu, Gao Liu, Chengqiang Lu, Keming Lu, Jianxin Ma, Rui Men, Xingzhang Ren, Xuancheng Ren, Chuanqi Tan, Sinan Tan, Jianhong Tu, Peng Wang, Shijie Wang, Wei Wang, Shengguang Wu, Benfeng Xu, Jin Xu, An~Yang, Hao Yang, Jian Yang, Shusheng Yang, Yang Yao, Bowen Yu, Hongyi Yuan, Zheng Yuan, Jianwei Zhang, Xingxuan Zhang, Yichang Zhang, Zhenru Zhang, Chang Zhou, Jingren Zhou, Xiaohuan Zhou, and Tianhang Zhu.
\newblock Qwen technical report.
\newblock \emph{CoRR}, abs/2309.16609, 2023{\natexlab{a}}.
\newblock \doi{10.48550/ARXIV.2309.16609}.
\newblock URL \url{https://doi.org/10.48550/arXiv.2309.16609}.

\bibitem[Bai et~al.(2023{\natexlab{b}})Bai, Lv, Zhang, Lyu, Tang, Huang, Du, Liu, Zeng, Hou, Dong, Tang, and Li]{bai2023longbench}
Yushi Bai, Xin Lv, Jiajie Zhang, Hongchang Lyu, Jiankai Tang, Zhidian Huang, Zhengxiao Du, Xiao Liu, Aohan Zeng, Lei Hou, Yuxiao Dong, Jie Tang, and Juanzi Li.
\newblock Longbench: {A} bilingual, multitask benchmark for long context understanding.
\newblock \emph{CoRR}, abs/2308.14508, 2023{\natexlab{b}}.
\newblock \doi{10.48550/ARXIV.2308.14508}.
\newblock URL \url{https://doi.org/10.48550/arXiv.2308.14508}.

\bibitem[Cai et~al.(2024)Cai, Cao, Chen, Chen, Chen, Chen, Chen, Chen, Chen, Chu, Dong, Duan, Fan, Fei, Gao, Ge, Gu, Gu, Gui, Guo, Guo, He, Hu, Huang, Jiang, Jiao, Jin, Lei, Li, Li, Li, Li, Li, Li, Liu, Liu, Hong, Liu, Liu, Liu, Lv, Lv, Lv, Ma, Ma, Ma, Ning, Ouyang, Qiu, Qu, Shang, Shao, Song, Song, Sui, Sun, Sun, Tang, Wang, Wang, Wang, Wang, Wang, Wang, Wang, Wei, Weng, Wu, Xiong, and et~al.]{2023internlm}
Zheng Cai, Maosong Cao, Haojiong Chen, Kai Chen, Keyu Chen, Xin Chen, Xun Chen, Zehui Chen, Zhi Chen, Pei Chu, Xiaoyi Dong, Haodong Duan, Qi~Fan, Zhaoye Fei, Yang Gao, Jiaye Ge, Chenya Gu, Yuzhe Gu, Tao Gui, Aijia Guo, Qipeng Guo, Conghui He, Yingfan Hu, Ting Huang, Tao Jiang, Penglong Jiao, Zhenjiang Jin, Zhikai Lei, Jiaxing Li, Jingwen Li, Linyang Li, Shuaibin Li, Wei Li, Yining Li, Hongwei Liu, Jiangning Liu, Jiawei Hong, Kaiwen Liu, Kuikun Liu, Xiaoran Liu, Chengqi Lv, Haijun Lv, Kai Lv, Li~Ma, Runyuan Ma, Zerun Ma, Wenchang Ning, Linke Ouyang, Jiantao Qiu, Yuan Qu, Fukai Shang, Yunfan Shao, Demin Song, Zifan Song, Zhihao Sui, Peng Sun, Yu~Sun, Huanze Tang, Bin Wang, Guoteng Wang, Jiaqi Wang, Jiayu Wang, Rui Wang, Yudong Wang, Ziyi Wang, Xingjian Wei, Qizhen Weng, Fan Wu, Yingtong Xiong, and et~al.
\newblock Internlm2 technical report.
\newblock \emph{CoRR}, abs/2403.17297, 2024.
\newblock \doi{10.48550/ARXIV.2403.17297}.
\newblock URL \url{https://doi.org/10.48550/arXiv.2403.17297}.

\bibitem[Del \& Fishel(2023)Del and Fishel]{del2023true}
Maksym Del and Mark Fishel.
\newblock True detective: A deep abductive reasoning benchmark undoable for gpt-3 and challenging for gpt-4.
\newblock In \emph{Proceedings of the 12th Joint Conference on Lexical and Computational Semantics (* SEM 2023)}, pp.\  314--322, 2023.

\bibitem[Dubey et~al.(2024)Dubey, Jauhri, Pandey, Kadian, Al-Dahle, Letman, Mathur, Schelten, Yang, Fan, et~al.]{dubey2024llama}
Abhimanyu Dubey, Abhinav Jauhri, Abhinav Pandey, Abhishek Kadian, Ahmad Al-Dahle, Aiesha Letman, Akhil Mathur, Alan Schelten, Amy Yang, Angela Fan, et~al.
\newblock The llama 3 herd of models.
\newblock \emph{arXiv preprint arXiv:2407.21783}, 2024.

\bibitem[Gu et~al.(2024)Gu, Zhang, Zhu, Chen, Huang, Zhang, Wang, Ye, Gao, Feng, and Xiao]{gu2024detectbenchlargelanguagemodel}
Zhouhong Gu, Lin Zhang, Xiaoxuan Zhu, Jiangjie Chen, Wenhao Huang, Yikai Zhang, Shusen Wang, Zheyu Ye, Yan Gao, Hongwei Feng, and Yanghua Xiao.
\newblock Detectbench: Can large language model detect and piece together implicit evidence?, 2024.
\newblock URL \url{https://arxiv.org/abs/2406.12641}.

\bibitem[Hendrycks et~al.(2021)Hendrycks, Burns, Basart, Zou, Mazeika, Song, and Steinhardt]{hendrycks2020MMLU}
Dan Hendrycks, Collin Burns, Steven Basart, Andy Zou, Mantas Mazeika, Dawn Song, and Jacob Steinhardt.
\newblock Measuring massive multitask language understanding.
\newblock In \emph{9th International Conference on Learning Representations, {ICLR} 2021, Virtual Event, Austria, May 3-7, 2021}. OpenReview.net, 2021.
\newblock URL \url{https://openreview.net/forum?id=d7KBjmI3GmQ}.

\bibitem[Huang et~al.(2023)Huang, Bai, Zhu, Zhang, Zhang, Su, Liu, Lv, Zhang, Lei, Fu, Sun, and He]{huang2023CEVAL}
Yuzhen Huang, Yuzhuo Bai, Zhihao Zhu, Junlei Zhang, Jinghan Zhang, Tangjun Su, Junteng Liu, Chuancheng Lv, Yikai Zhang, Jiayi Lei, Yao Fu, Maosong Sun, and Junxian He.
\newblock C-eval: {A} multi-level multi-discipline chinese evaluation suite for foundation models.
\newblock In Alice Oh, Tristan Naumann, Amir Globerson, Kate Saenko, Moritz Hardt, and Sergey Levine (eds.), \emph{Advances in Neural Information Processing Systems 36: Annual Conference on Neural Information Processing Systems 2023, NeurIPS 2023, New Orleans, LA, USA, December 10 - 16, 2023}, 2023.
\newblock URL \url{http://papers.nips.cc/paper\_files/paper/2023/hash/c6ec1844bec96d6d32ae95ae694e23d8-Abstract-Datasets\_and\_Benchmarks.html}.

\bibitem[InternLM(2024)]{InternLM25}
InternLM.
\newblock Internlm2.5-7b, July 2024.
\newblock URL \url{https://huggingface.co/internlm/internlm2_5-7b}.

\bibitem[Kamradt(2023)]{haystack}
Greg Kamradt.
\newblock Needle in a haystack - pressure testing llms, 2023.
\newblock URL \url{https://github.com/gkamradt/ LLMTest_NeedleInAHaystack}.

\bibitem[Karpinska et~al.(2024)Karpinska, Thai, Lo, Goyal, and Iyyer]{karpinska2024one}
Marzena Karpinska, Katherine Thai, Kyle Lo, Tanya Goyal, and Mohit Iyyer.
\newblock One thousand and one pairs: A" novel" challenge for long-context language models.
\newblock \emph{arXiv preprint arXiv:2406.16264}, 2024.

\bibitem[Kocisk{\'{y}} et~al.(2018)Kocisk{\'{y}}, Schwarz, Blunsom, Dyer, Hermann, Melis, and Grefenstette]{kovcisky2018narrativeqa}
Tom{\'{a}}s Kocisk{\'{y}}, Jonathan Schwarz, Phil Blunsom, Chris Dyer, Karl~Moritz Hermann, G{\'{a}}bor Melis, and Edward Grefenstette.
\newblock The narrativeqa reading comprehension challenge.
\newblock \emph{Trans. Assoc. Comput. Linguistics}, 6:\penalty0 317--328, 2018.
\newblock \doi{10.1162/TACL\_A\_00023}.
\newblock URL \url{https://doi.org/10.1162/tacl\_a\_00023}.

\bibitem[Kuratov et~al.(2024)Kuratov, Bulatov, Anokhin, Rodkin, Sorokin, Sorokin, and Burtsev]{kuratov2024babilong}
Yuri Kuratov, Aydar Bulatov, Petr Anokhin, Ivan Rodkin, Dmitry Sorokin, Artyom Sorokin, and Mikhail Burtsev.
\newblock Babilong: Testing the limits of llms with long context reasoning-in-a-haystack.
\newblock \emph{arXiv preprint arXiv:2406.10149}, 2024.

\bibitem[Li et~al.(2023)Li, Wang, Zheng, and Zhang]{li2023loogle}
Jiaqi Li, Mengmeng Wang, Zilong Zheng, and Muhan Zhang.
\newblock Loogle: Can long-context language models understand long contexts?
\newblock \emph{CoRR}, abs/2311.04939, 2023.
\newblock \doi{10.48550/ARXIV.2311.04939}.
\newblock URL \url{https://doi.org/10.48550/arXiv.2311.04939}.

\bibitem[Li et~al.(2024)Li, Zhang, Liu, and Chen]{li2024needlebench}
Mo~Li, Songyang Zhang, Yunxin Liu, and Kai Chen.
\newblock Needlebench: Can llms do retrieval and reasoning in 1 million context window?
\newblock \emph{arXiv preprint arXiv:2407.11963}, 2024.

\bibitem[Lin(2004)]{lin2004rouge}
Chin-Yew Lin.
\newblock Rouge: A package for automatic evaluation of summaries.
\newblock In \emph{Text summarization branches out}, pp.\  74--81, 2004.

\bibitem[Liu et~al.(2023)Liu, Iter, Xu, Wang, Xu, and Zhu]{liu2023g}
Yang Liu, Dan Iter, Yichong Xu, Shuohang Wang, Ruochen Xu, and Chenguang Zhu.
\newblock G-eval: Nlg evaluation using gpt-4 with better human alignment.
\newblock In \emph{Proceedings of the 2023 Conference on Empirical Methods in Natural Language Processing}, pp.\  2511--2522, 2023.

\bibitem[Mondorf \& Plank(2024)Mondorf and Plank]{mondorf2024beyond}
Philipp Mondorf and Barbara Plank.
\newblock Beyond accuracy: Evaluating the reasoning behavior of large language models--a survey.
\newblock \emph{arXiv preprint arXiv:2404.01869}, 2024.

\bibitem[OpenAI(2023)]{achiam2023gpt4}
OpenAI.
\newblock {GPT-4} technical report.
\newblock \emph{CoRR}, abs/2303.08774, 2023.
\newblock \doi{10.48550/ARXIV.2303.08774}.
\newblock URL \url{https://doi.org/10.48550/arXiv.2303.08774}.

\bibitem[OpenAI(2024)]{openai2024o1}
OpenAI.
\newblock Introducing openai o1, 2024.
\newblock URL \url{https://openai.com/o1/}.

\bibitem[Saito(2007)]{saito2007orthodox}
Satomi Saito.
\newblock Culture and authenticity : the discursive space of japanese detective fiction and the formation of the national imaginary.
\newblock 2007.
\newblock URL \url{https://api.semanticscholar.org/CorpusID:190048951}.

\bibitem[Sprague et~al.(2024)Sprague, Ye, Bostrom, Chaudhuri, and Durrett]{sprague2024musrtestinglimitschainofthought}
Zayne Sprague, Xi~Ye, Kaj Bostrom, Swarat Chaudhuri, and Greg Durrett.
\newblock Musr: Testing the limits of chain-of-thought with multistep soft reasoning, 2024.
\newblock URL \url{https://arxiv.org/abs/2310.16049}.

\bibitem[Sun et~al.(2024)Sun, Zhang, He, Li, Cheng, Liu, Yan, Shao, Tang, Zhang, Zhao, Chen, Zheng, Zhou, Li, Zhan, Zhou, Li, Yang, Wu, Yin, Huang, Jiang, and Qiu]{Sun2024moss}
Tianxiang Sun, Xiaotian Zhang, Zhengfu He, Peng Li, Qinyuan Cheng, Xiangyang Liu, Hang Yan, Yunfan Shao, Qiong Tang, Shiduo Zhang, Xingjian Zhao, Ke~Chen, Yining Zheng, Zhejian Zhou, Ruixiao Li, Jun Zhan, Yunhua Zhou, Linyang Li, Xiaogui Yang, Lingling Wu, Zhangyue Yin, Xuanjing Huang, Yu-Gang Jiang, and Xipeng Qiu.
\newblock Moss: An open conversational large language model.
\newblock \emph{Machine Intelligence Research}, 2024.
\newblock ISSN 2731-5398.
\newblock \doi{10.1007/s11633-024-1502-8}.
\newblock URL \url{https://doi.org/10.1007/s11633-024-1502-8}.

\bibitem[Team(2024)]{qwen25}
Qwen Team.
\newblock Qwen2.5: A party of foundation models, September 2024.
\newblock URL \url{https://qwenlm.github.io/blog/qwen2.5/}.

\bibitem[Touvron et~al.(2023)Touvron, Martin, Stone, Albert, Almahairi, Babaei, Bashlykov, Batra, Bhargava, Bhosale, Bikel, Blecher, Canton{-}Ferrer, Chen, Cucurull, Esiobu, Fernandes, Fu, Fu, Fuller, Gao, Goswami, Goyal, Hartshorn, Hosseini, Hou, Inan, Kardas, Kerkez, Khabsa, Kloumann, Korenev, Koura, Lachaux, Lavril, Lee, Liskovich, Lu, Mao, Martinet, Mihaylov, Mishra, Molybog, Nie, Poulton, Reizenstein, Rungta, Saladi, Schelten, Silva, Smith, Subramanian, Tan, Tang, Taylor, Williams, Kuan, Xu, Yan, Zarov, Zhang, Fan, Kambadur, Narang, Rodriguez, Stojnic, Edunov, and Scialom]{Touvron2023llama2}
Hugo Touvron, Louis Martin, Kevin Stone, Peter Albert, Amjad Almahairi, Yasmine Babaei, Nikolay Bashlykov, Soumya Batra, Prajjwal Bhargava, Shruti Bhosale, Dan Bikel, Lukas Blecher, Cristian Canton{-}Ferrer, Moya Chen, Guillem Cucurull, David Esiobu, Jude Fernandes, Jeremy Fu, Wenyin Fu, Brian Fuller, Cynthia Gao, Vedanuj Goswami, Naman Goyal, Anthony Hartshorn, Saghar Hosseini, Rui Hou, Hakan Inan, Marcin Kardas, Viktor Kerkez, Madian Khabsa, Isabel Kloumann, Artem Korenev, Punit~Singh Koura, Marie{-}Anne Lachaux, Thibaut Lavril, Jenya Lee, Diana Liskovich, Yinghai Lu, Yuning Mao, Xavier Martinet, Todor Mihaylov, Pushkar Mishra, Igor Molybog, Yixin Nie, Andrew Poulton, Jeremy Reizenstein, Rashi Rungta, Kalyan Saladi, Alan Schelten, Ruan Silva, Eric~Michael Smith, Ranjan Subramanian, Xiaoqing~Ellen Tan, Binh Tang, Ross Taylor, Adina Williams, Jian~Xiang Kuan, Puxin Xu, Zheng Yan, Iliyan Zarov, Yuchen Zhang, Angela Fan, Melanie Kambadur, Sharan Narang, Aur{\'{e}}lien Rodriguez, Robert Stojnic, Sergey Edunov,
  and Thomas Scialom.
\newblock Llama 2: Open foundation and fine-tuned chat models.
\newblock \emph{CoRR}, abs/2307.09288, 2023.
\newblock \doi{10.48550/ARXIV.2307.09288}.
\newblock URL \url{https://doi.org/10.48550/arXiv.2307.09288}.

\bibitem[Wang et~al.(2024)Wang, Ning, Pan, Wu, Guo, Deng, Bao, Wang, and Zhang]{wang2024novelqa}
Cunxiang Wang, Ruoxi Ning, Boqi Pan, Tonghui Wu, Qipeng Guo, Cheng Deng, Guangsheng Bao, Qian Wang, and Yue Zhang.
\newblock Novelqa: {A} benchmark for long-range novel question answering.
\newblock \emph{CoRR}, abs/2403.12766, 2024.
\newblock \doi{10.48550/ARXIV.2403.12766}.
\newblock URL \url{https://doi.org/10.48550/arXiv.2403.12766}.

\bibitem[Wu et~al.(2021)Wu, Ouyang, Ziegler, Stiennon, Lowe, Leike, and Christiano]{wu2021recursively}
Jeff Wu, Long Ouyang, Daniel~M. Ziegler, Nisan Stiennon, Ryan Lowe, Jan Leike, and Paul~F. Christiano.
\newblock Recursively summarizing books with human feedback.
\newblock \emph{CoRR}, abs/2109.10862, 2021.
\newblock URL \url{https://arxiv.org/abs/2109.10862}.

\bibitem[Xia et~al.(2024)Xia, Li, Liu, Wu, and Liu]{xia2024evaluating}
Shijie Xia, Xuefeng Li, Yixin Liu, Tongshuang Wu, and Pengfei Liu.
\newblock Evaluating mathematical reasoning beyond accuracy.
\newblock \emph{arXiv preprint arXiv:2404.05692}, 2024.

\bibitem[Xu et~al.(2023)Xu, Ping, Wu, McAfee, Zhu, Liu, Subramanian, Bakhturina, Shoeybi, and Catanzaro]{xu2023ragvslong}
Peng Xu, Wei Ping, Xianchao Wu, Lawrence McAfee, Chen Zhu, Zihan Liu, Sandeep Subramanian, Evelina Bakhturina, Mohammad Shoeybi, and Bryan Catanzaro.
\newblock Retrieval meets long context large language models.
\newblock \emph{CoRR}, abs/2310.03025, 2023.
\newblock \doi{10.48550/ARXIV.2310.03025}.
\newblock URL \url{https://doi.org/10.48550/arXiv.2310.03025}.

\bibitem[Yang et~al.(2018)Yang, Qi, Zhang, Bengio, Cohen, Salakhutdinov, and Manning]{yang2018hotpotqa}
Zhilin Yang, Peng Qi, Saizheng Zhang, Yoshua Bengio, William~W. Cohen, Ruslan Salakhutdinov, and Christopher~D. Manning.
\newblock Hotpotqa: {A} dataset for diverse, explainable multi-hop question answering.
\newblock In Ellen Riloff, David Chiang, Julia Hockenmaier, and Jun'ichi Tsujii (eds.), \emph{Proceedings of the 2018 Conference on Empirical Methods in Natural Language Processing, Brussels, Belgium, October 31 - November 4, 2018}, pp.\  2369--2380. Association for Computational Linguistics, 2018.
\newblock \doi{10.18653/V1/D18-1259}.
\newblock URL \url{https://doi.org/10.18653/v1/d18-1259}.

\bibitem[Zeng et~al.(2023{\natexlab{a}})Zeng, Liu, Du, Wang, Lai, Ding, Yang, Xu, Zheng, Xia, Tam, Ma, Xue, Zhai, Chen, Liu, Zhang, Dong, and Tang]{zeng2022glm}
Aohan Zeng, Xiao Liu, Zhengxiao Du, Zihan Wang, Hanyu Lai, Ming Ding, Zhuoyi Yang, Yifan Xu, Wendi Zheng, Xiao Xia, Weng~Lam Tam, Zixuan Ma, Yufei Xue, Jidong Zhai, Wenguang Chen, Zhiyuan Liu, Peng Zhang, Yuxiao Dong, and Jie Tang.
\newblock {GLM-130B:} an open bilingual pre-trained model.
\newblock In \emph{The Eleventh International Conference on Learning Representations, {ICLR} 2023, Kigali, Rwanda, May 1-5, 2023}. OpenReview.net, 2023{\natexlab{a}}.
\newblock URL \url{https://openreview.net/pdf?id=-Aw0rrrPUF}.

\bibitem[Zeng et~al.(2023{\natexlab{b}})Zeng, Liu, Du, Wang, Lai, Ding, Yang, Xu, Zheng, Xia, Tam, Ma, Xue, Zhai, Chen, Liu, Zhang, Dong, and Tang]{zeng2023chatglm}
Aohan Zeng, Xiao Liu, Zhengxiao Du, Zihan Wang, Hanyu Lai, Ming Ding, Zhuoyi Yang, Yifan Xu, Wendi Zheng, Xiao Xia, Weng~Lam Tam, Zixuan Ma, Yufei Xue, Jidong Zhai, Wenguang Chen, Zhiyuan Liu, Peng Zhang, Yuxiao Dong, and Jie Tang.
\newblock {GLM-130B:} an open bilingual pre-trained model.
\newblock In \emph{The Eleventh International Conference on Learning Representations, {ICLR} 2023, Kigali, Rwanda, May 1-5, 2023}. OpenReview.net, 2023{\natexlab{b}}.
\newblock URL \url{https://openreview.net/pdf?id=-Aw0rrrPUF}.

\bibitem[Zhang et~al.(2024)Zhang, Chen, Hu, Xu, Chen, Hao, Han, Thai, Wang, Liu, and Sun]{zhang2023infinitebench}
Xinrong Zhang, Yingfa Chen, Shengding Hu, Zihang Xu, Junhao Chen, Moo~Khai Hao, Xu~Han, Zhen~Leng Thai, Shuo Wang, Zhiyuan Liu, and Maosong Sun.
\newblock {\(\infty\)}bench: Extending long context evaluation beyond 100k tokens.
\newblock \emph{CoRR}, abs/2402.13718, 2024.
\newblock \doi{10.48550/ARXIV.2402.13718}.
\newblock URL \url{https://doi.org/10.48550/arXiv.2402.13718}.

\end{thebibliography}
\bibliographystyle{iclr2025_conference}

\newpage
\appendix

\section{Annotation Guideline}\label{sec:guideline}

\paragraph{Data format.} 
For each novel, we require the annotators to annotate (1) several multiple-choice questions involving reasoning with (2) the answers to the question and (3) the multi-step reasoning chains.
For each step in reasoning chains, we annotate (4) tags indicating whether the step is a piece of evidence in the original novel and the corresponding position.
An example of our annotated data is in Figure~\ref{fig:exemplery-annotated-data}.

\begin{figure}[!t]
    \framebox{
    \parbox{0.45\textwidth}{
    \small
    \{\newline
    \textbf{"question":}"Which of the following is the reason for the disappearance of Sainsbury Seale?", \newline
    
    \textbf{"options":} \newline
                "A": "left voluntarily.",\newline
                "B": "met an untimely end.",\newline
                "C": "eloped with someone.",\newline
                "D": "Sudden memory loss." \newline

    \textbf{"answer":}"B",\newline
    
    \textbf{"reasoning":} [\newline
    \evidence{"Ms Sainsbury-Seal did not take her luggage with her when she disappeared.",}\newline
    \inference{"This does not appear to be a voluntary departure.",}\newline
    \evidence{"Ms Seale had a dinner date with a friend to play solitaire.",}\newline
    \inference{"Normally at the appointed time she would have been back at the hotel.",}\newline
    \inference{"Therefore, based on the above evidences, it is surmised that it was Sainsbury Seale who met an untimely end."}\smallskip\newline
    ],\newline
    
    \textbf{"evidence\_position":}[\evidence{740},\inference{-1},\evidence{734},\inference{-1},\inference{-1}],
    \smallskip\newline
    
    \textbf{"answer\_position":} 1202\newline
    \}

    }
    }
    \caption{An example of a multiple-choice annotation in DetectiveQA. We highlight the evidences of reasoning in \evidence{blue italics}, and inference in \inference{green plain typeface} . the "reasoning" part includes evidences and inferences, while in the "evidence\_position" field, the part corresponding to the evidence will be the paragraph in which the evidence occurs in the article, while the part corresponding to the reference will be -1.} 
    \label{fig:exemplery-annotated-data}
\end{figure}
\begin{figure*}[h] 
    \includegraphics[width=1.\textwidth]{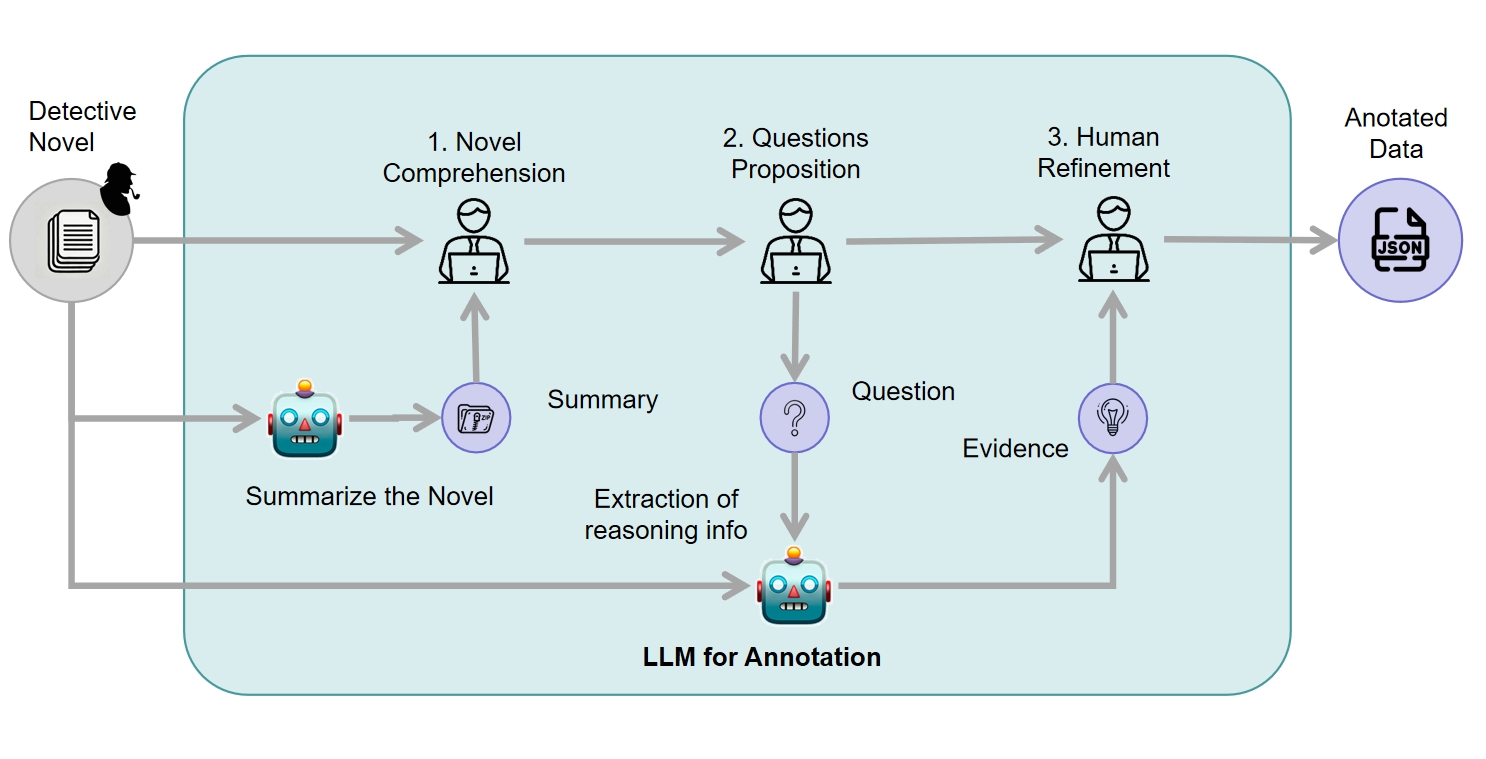}
    \caption{Annotation process}
    \label{fig: acceleration}
\end{figure*}

\paragraph{Annotation procedure.}
We visualize the annotation process in Figure~\ref{fig: acceleration}. 
With the collected novels, we first apply our agents to summarize the novels, locating rationales from the detective in the novel and proposing candidate questions for each rationale.
The prompts we use are in Fig.~\ref{fig:exemplery-annotated-data}

With the outputs from the agent, we require the annotators to refine the data with the following procedure.
\begin{enumerate}[nosep,label=(\roman*)]
\item \textbf{On questions.} The annotator should determine whether a question requires reasoning to answer. This involves assessing if answering the question necessitates extracting and synthesizing clues from the original text. Questions that can be answered through simple retrieval of a specific answer from the text, without requiring any additional contextual information, should be filtered out. Additionally, the validity of the question should be evaluated, with manual modifications applied to the original question as needed.
\item \textbf{On options.} The question must be accompanied by four options, including the correct answer and a minimum of one distractor.
\item \textbf{On evidence.} The annotators evaluate whether the model-provided evidence is relevant to the question, complete, and accurate. 
They should eliminate incorrect or irrelevant evidence and supplement missing ones. 
\item \textbf{On answers and rationales.}
The rationales should be separated into steps with each step being an evidence or inferences.
\item \textbf{On positions.} The annotators should ensure the answer positions and the evidence positions are correct.
\end{enumerate}

\paragraph{Agent Workflow}\label{sec:agent_workflow}

Figure~\ref{fig: acceleration} illustrates the specific process of accelerated annotation through agent workflow. LLM extracts a wide variety of key information, which a human then integrates and constructs into a complete annotation file for the reasoning problem.

\section{Details in Reasoning Metric}\label{sec:gpt_score}
In our step-wise reasoning metric, we mentioned the use of GPT4 for the number of contained leads, and we used the following prompt template~\ref{fig: prompt template}.We use LLMs such as Kimi, Claude3, and others.

\begin{figure}[h]
    \framebox{
    \parbox{1.0\textwidth}{
    \small
    In order to evaluate a question-answering (QA) system's reasoning process regarding a particular inference question, specifically whether the reasoning process correctly includes certain reference steps, multiple reference steps will be provided.\newline

    Due to the nature of the inference question being based on a detective novel, the reasoning process may involve some sensitive content. However, for the purpose of this evaluation, please focus solely on determining whether the reasoning process explicitly or implicitly includes the provided reference steps.\newline
    
    The QA system's output for the reasoning process may not explicitly mention the provided reference steps, but it might implicitly incorporate them. In such cases, it should still be considered as correct including the provided reference steps.\newline
    
    The reasoning process output by the QA system and the reference reference steps to be considered are presented below. Please objectively assess whether the QA system's reasoning process explicitly or implicitly includes the provided reference steps, and clearly state which reference steps are included.\newline
    \newline
    Reasoning Process:\newline
    \textbf{[Reasoning Process]}\newline
    \newline
    reference steps:\newline
    \textbf{[reference steps]}\newline
    \newline
    Provide an initial sentence explaining whether the reasoning process explicitly or implicitly includes each reasoning step. Then, in the second line, specify the indices of the included reference steps in a list format, such as [0, 1, 2, 3, ...].\newline
    \newline
    Your response should maintain this format:\newline
    \newline
    Explanation: <One-sentence explanation>\newline
    Included Reference Steps: [Indices of the included reference steps]\newline

    }
    }
    \caption{GPT-4 Questioning Template: Replace \textbf{bolded font} with evidence and Model's Inference Process in Query.}
    \label{fig: prompt template}
\end{figure} 
This template asks enough questions to get usable responses without adding additional samples for a few shots to help answer.

\section{More Validation Results}\label{sec:more_valid}
Table ~\ref{tab: part statistic} lists the statistics of the novel parts that were manual annotation and accelerated through agent workflow-assisted annotation. While there is an increase in the length of questions for agent workflow-assisted accelerated annotation, the evidence and reasoning sections of manual annotation are more detailed and have slightly less coverage than agent workflow-assisted accelerated annotation. Overall, the difference in quality between manual and agent workflow-assisted accelerated annotation was minimal, suggesting that the use of agent workflow-assisted accelerated annotation is feasible.

\begin{table}[t]
\centering
\small
\resizebox{0.48\textwidth}{!}{
\begin{tabular}{lcc}
\toprule
\textbf{Statistic} & \textbf{w/ agent workflow} & \textbf{w/o agent workflow} \\ 
 & \textbf{(Max/Min/Avg)} & \textbf{(Max/Min/Avg)} \\
\midrule
context length & 148k/4k/81k & 167k/4k/94k \\
% Coverage factor & 95.6/ 0.3/ 35.4 & 99.0/ 0.2/ 44.6 \\
% evidence & 17 / 2 / 5.98 & 21 / 2 / 5.82 \\
% inference step &5 / 1 / 1.82 & 5 / 1 / 1.82 \\
% evidence length & 740 / 37 / 196.29 & 404 / 30 / 150.09 \\
reference steps & 26 / 4 / 7.65 & 21 / 3 / 7.80 \\
%inference length & 357 / 23 / 105.64 & 244 / 16 / 101.05 \\
reasoning length & 606 / 78 / 251.14 & 1073 / 92 / 301.94 \\
total questions &  308 & 338 \\
\bottomrule
\end{tabular}}

\caption{Annotation w/ agent workflow vs. Annotation w/o agent workflow Statistics Comparison
where context length refers to the length of the problem and the meaning of the remaining metrics is detailed in Section~\ref{sec:valid}.
}
\label{tab: part statistic}
\end{table}

\begin{figure}[t]
    \centering
    \includegraphics[width=.4\textwidth]{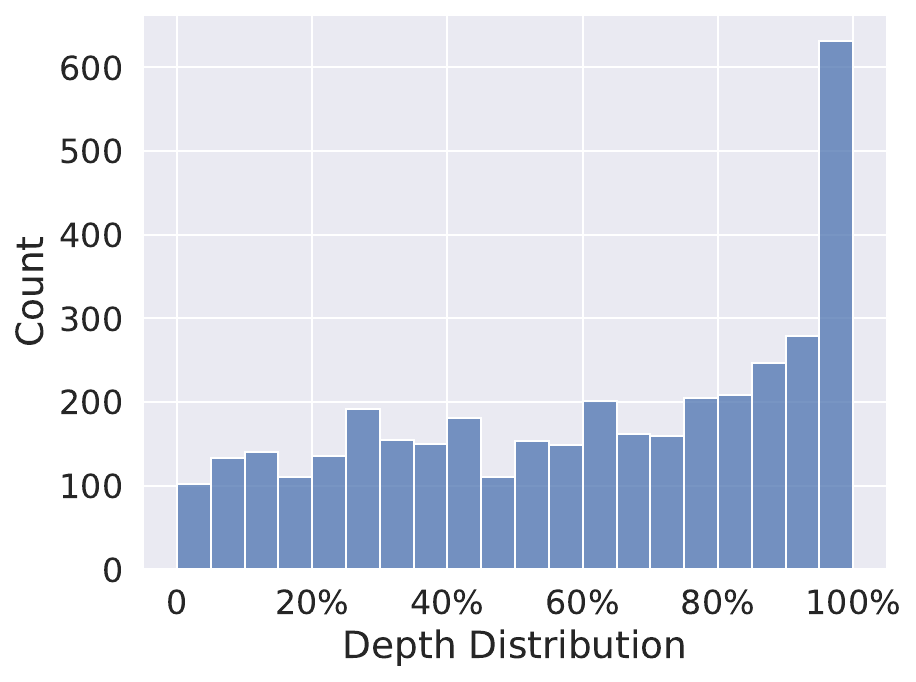}
    \caption{Distribution of different evidence depths. For each piece of evidence, we take as depth the percentage of the total number of words from the beginning of the word in which it occurs
    % \textcolor{red}{TODOs: describe how to depths are calculated}
    }
    \label{fig:depth distribution}
\end{figure}

\paragraph{Depth Distribution}\label{sec:depth distribution}
The exact distribution can be seen in Figure~\ref{fig:depth distribution}, where there will be more clues at 100\% depth since the longer dependent questions will also have valid clues in the closer locations.

\section{More Evaluation Results}\label{sec:more_eval}
We have done quite a lot of experiments under the 32K input length model, and we can find that the experimental results of the 32K model are all relatively unsatisfactory.
The experimental results are shown in Table~\ref{tab:32k results}

However, a larger number of parameters would allow the model to make full use of the information in the 32K text and its own reasoning power to mitigate the problem.

% \section{Agent Workflow}\label{sec:agent_workflow}

% \begin{figure*}[h] 
%     \includegraphics[width=1.\textwidth]{figures/Acceleration.png}
%     \caption{Annotation process}
%     \label{fig: acceleration}
% \end{figure*}

% \subsection{Agent Workflow-assisted Annotating Process}
% Figure~\ref{fig: acceleration} illustrates the specific process of accelerated annotation through agent workflow. LLM extracts a wide variety of key information, which a human then integrates and constructs into a complete annotation file for the reasoning problem.

% \subsection{Annotating Specification}

\begin{table*}[t]
\centering
\caption{Performance of models supporting context lengths of 32K or less }
\resizebox{0.98\linewidth}{!}{
\begin{tabular}{lccccccc}  
\toprule
\multirow{2}{*}{\textbf{Models}} & \multicolumn{3}{c}{\textbf{Question+Context}} & \multicolumn{3}{c}{\textbf{Question-Only}} & \multirow{2}{*}{\textbf{Win Rate}}\\ 
& Answer & Reasoning & G.M. & Answer & Reasoning & G.M. & \\
\midrule
% \midrule
LongChat-v1.5-7B-32k & 29.33 & 11.07 & 18.01 & 27.83 & 5.72 & 12.59 & 33.58 \\ 
Vicuna-v1.5-7B-16k & 30.33 & 12.63 & 19.57 & 27.67 & 6.69 & 13.60 & 32.57 \\ 
Qwen1.5-7B-8k & 49.50 & 10.09 & 22.34 & 35.33 & 7.74 & 16.53 & 60.71\\ 
Qwen1.5-72B-32K & \textbf{70.67} & 19.69 & \textbf{37.30} & \textbf{44.67} & 10.55 & \textbf{21.70}	& 76.51 \\
\bottomrule
\end{tabular} 
}

\label{tab:32k results}
\end{table*}

\end{document}